\pdfoutput=1
\documentclass{article}
\usepackage{arxiv}
\usepackage[utf8]{inputenc} 
\usepackage[T1]{fontenc}    
\usepackage{hyperref}       
\usepackage{amsmath,amssymb,bm}
\usepackage{cleveref}       
\usepackage{url}            
\usepackage{booktabs}       
\usepackage{amsfonts}       
\usepackage{natbib}         
\usepackage{nicefrac}       
\usepackage{microtype}      
\usepackage{graphicx}
\usepackage{xcolor}
\usepackage{lmodern}
\usepackage{tikz}
\usetikzlibrary{positioning,calc}
\graphicspath{{Figures/}}

\newcommand{\R}{\mathbb{R}}

\newcommand{\bA}{\mathbf{A}}

\newcommand{\bx}{\mathbf{x}}

\title{\LARGE Conformalized Physics-Informed Weak Regression for Parameter Estimation in Epidemic Models}
\author{
Allan P. Engsig-Karup \\
Department of Applied Mathematics and Computer Science \\
Technical University of Denmark \\
Kongens Lyngby, Denmark \\
\texttt{apek@dtu.dk} \\
}

\begin{document}
\maketitle

\begin{abstract}
Compartment models of infectious diseases are often linear in a vector of coefficients, each of which is a physical parameter or a product or ratio of parameters. We
exploit this by using closed-form least squares, and equip the estimates with
distribution-free uncertainty. Physics-informed regression (PIR) fits the parameters to finite-difference estimates of
the time derivatives. We introduce a new physics-informed weak regression (PIWR) method, which integrates the model in time instead,
contains no derivatives of the data, estimates the initial state jointly, and is high-order accurate with cubic-spline
quadrature. To quantify uncertainty we wrap both estimators in split conformal prediction. The calibration population
is generated by the physics-based model itself: epidemics with known parameters are simulated, perturbed with the data's
noise model and processed by the same fixed estimator. This yields intervals for all parameters of a model at once
with a finite-sample coverage guarantee. A residual-adaptive factor, computed without ground truth, scales the uncertainty intervals
of each data set. With PIWR we call the new conformalized approach PIWR-RACC. On all eleven benchmark models of disease-informed neural networks (DINNs), from a four-state SIDR model of COVID-19 to a nine-state host--vector model of Zika, PIWR is the most accurate
estimator in 15 of 22 model/noise settings. Where the sampling resolves the dynamics, PIWR or sixth-order finite-difference-based PIR-FD6 is one to four orders of magnitude more accurate than the reported DINN results, and each fit takes about a millisecond, against minutes to hours of training of DINN to perform the same parameter estimation. The simultaneous 90\,\% intervals achieve 87--92\,\% empirical coverage in all 88 tested settings. They are narrow where the estimates are accurate and become wide exactly where the data do not resolve the dynamics. On the seven-compartment S3I3R model of the original PIR study, PIWR lowers the reported PIR errors by about three and the DINN errors by about five orders of magnitude. On real data from the 1978 boarding-school influenza outbreak, a variable-projection variant of PIWR gives calibrated 95\,\% intervals that overlap the credible intervals of a Bayesian analysis. On a time-varying transmission benchmark from a recent PINN study, a B-spline expansion of $\beta(t)$ keeps PIR and PIWR linear. They are more accurate than the PINNs in three of four cases, mostly by one to two orders of magnitude, and at least comparable in the noisiest one, in milliseconds instead of tens of minutes, and the conformal layer gives simultaneous bands for $\beta(t)$. For all benchmarks the complete conformal calibration that provides the uncertainty intervals takes less than two minutes.
\end{abstract}

\keywords{Parameter estimation \and Physics-informed regression \and Physics-Informed weak Regression \and Epidemic models \and Conformal prediction \and Uncertainty quantification}

\section{Introduction}\label{sec:introduction}

Compartment models are the standard language of mathematical epidemiology. They are systems of ordinary
differential equations (ODEs) whose coefficients for transmission, recovery, death and transition
rates are the quantities of interest. They determine reproduction numbers, peak sizes and the effect of
interventions. Estimating them from time series is an inverse problem, and over the last years
physics-informed neural networks (PINNs) \citep{RAISSI2019686,karniadakis2021physics,cuomo2022scientific} have
become a popular tool for it. An example is the family of \emph{disease-informed neural networks}
(DINNs) of \citet{Shaier2022DINN}, which identifies the parameters of eleven compartment models, from a
three-state SIDR model of COVID-19 to a nine-state host--vector model of Zika, by training a network for
each disease. It is reported at the time that training takes from 20 minutes to 91 hours per disease, the search range of every parameter must
be prescribed (typically $\pm 1$--$10\,\%$ around the literature value), and the estimates come without any
statement of their uncertainty which limits how much trust we can put into such estimates.

Almost all compartment models share a structure where their right-hand side is
\emph{linear in a vector of coefficients}, each of which is a physical parameter or a product of parameters (for
example $\beta_0\beta_1$ in a seasonally forced transmission rate $\beta_0(1+\beta_1\cos 2\pi t)$). The coefficients then
follow from a linear least-squares problem, and the physical parameters from an explicit map back. For such models, \citet{Nielsen2025PIR} proposed
\emph{physics-informed regression} (PIR) based on estimating the time derivatives of the states from the data, evaluate the
model library at the data, and solve one ordinary least-squares problem. PIR has the same physics-informed
ingredient as a PINN, namely, the residual of the governing equations, however, replaces training by a closed-form solve resulting in low computational cost and as a result achieve dramatic improvement over methods such as DINN derived from PINN for the problems considered. The accuracy of PIR is limited by the derivative estimate, which is the weak point whenever the data are coarse in time or noisy.

A point estimate of the parameters obtained through regression analysis is, however, only half of the answer. The estimated coefficients are
affected by measurement noise associate with data collection, which a derivative estimate amplifies, e.g. by the truncation error of the finite differences or of the quadrature, by an errors-in-variables bias, since the library is evaluated at noisy states,  by practical non-identifiability when the sampling does not resolve the dynamics, and by the inevitable mismatch between a compartment model and a real epidemic. The map $\bm\omega = h(\bm\theta)$ can amplify these errors further, for example through ratios of coefficients. Since quantities such as the reproduction number are used to decide on interventions, every estimate needs an honest statement of its uncertainty. Conventional approaches are well established. Nonlinear least squares provides an asymptotic covariance from the linearized model and an assumed
noise model \citep{BatesWatts1988}; profile likelihoods give confidence intervals and expose non-identifiable parameters \citep{Raue2009Profile}; parametric bootstrapping repeats the fit on resampled data \citep{Efron1979Bootstrap,Chowell2017Primer}; Bayesian inference by Markov chain Monte Carlo
\citep{Carpenter2017Stan,Grinsztajn2021Stan} or approximate Bayesian computation \citep{Toni2009ABC} yields posterior distributions, also for physics-informed neural networks \citep{yang2021bpinn}; model discrepancy can
be represented explicitly \citep{Kennedy2001Calibration}, and polynomial chaos expansions propagate parameter
uncertainty through epidemic models at a fraction of the cost of Monte Carlo sampling and quantify the sensitivity
of the predictions to each parameter \citep{Jensen2022PC}. These methods require repeated forward solves or
likelihood evaluations, and their intervals are valid only asymptotically or conditionally on the assumed prior
and noise model.

We propose to consider the perspective that {\em physics-informed regression offers a different route}. The least-squares covariance of PIR and PIWR is available at no cost, but it treats the residual as white noise and ignores the truncation bias and the correlation of the residuals, so it is not a trustworthy uncertainty on its own. Split conformal prediction \citep{vovk2005algorithmic,shafer2008tutorial,lei2018distribution,angelopoulos2023gentle} turns such a heuristic uncertainty into intervals with a finite-sample, distribution-free coverage guarantee, provided that calibration
examples with known truth are available. A physics-based model can supply them, because the model can be simulated numerically and the regression methods considered in this study costs about a millisecond. Hence, thousands of synthetic epidemics with known parameters can be passed through exactly the same estimator as the data. All error sources that the simulation contains, including noise, discretization and quadrature errors, the errors-in-variables bias and weak identifiability, are then calibrated jointly, without linearization and without a prior. The resulting guarantee is marginal over the calibration population rather than conditional on the observed data, and it does not cover model errors that the simulations do not contain. Conformalized physics-informed regression therefore complements rather than replaces likelihood-based and Bayesian analyses, and the two can be used together for validation, as we do for real data in \cref{sec:res:extra}.

\subsection{Contributions}
This paper makes three contributions related to developing new fast and reliable numerical parameter estimation techniques based on extending the idea of physics-informed regression for parameter-linear models \cite{Nielsen2025PIR}.

\textbf{Physics-informed weak regression (PIWR).} We integrate the model in time instead of differentiating
  the data. The integrated equations remain linear in the parameters, contain no derivatives of the data,
  and are solved jointly for the parameters and the unknown initial state. With cubic-spline quadrature the
  method is fourth-order accurate in the sampling interval. PIWR is the regression analogue of weak-form
  identification methods \citep{Varah1982,Ramsay2007,Messenger2021WSINDy}, here with a known model structure.

\textbf{Conformal prediction with physics-based models.} We wrap PIR and PIWR in a split conformal layer
  \citep{vovk2005algorithmic,shafer2008tutorial,lei2018distribution,angelopoulos2023gentle} that returns intervals
  for \emph{all} parameters of a model simultaneously, with a finite-sample, distribution-free coverage guarantee.
  A model coefficient has no natural calibration set, since every data set has exactly one true parameter
  vector. We use the physics-based model itself to generate one: a population of epidemics with known parameters,
  simulated and processed exactly like the data. The least-squares covariance sets the relative widths of the
  intervals, the calibration sets their common scale.

\textbf{Residual-adaptive conformal calibration (RACC).} We introduce a residual-adaptive factor that makes
  the uncertainty intervals of the parameter estimates depend on the individual data set. The relative residual of the regression measures how well the
  data satisfy the governing equations, grows when the sampling does not resolve the dynamics or the noise is
  large, and is available without ground truth. Scaling the conformal scores and intervals by a factor that
  increases with this residual widens the intervals for difficult data sets and narrows them for easy ones, while
  the finite-sample coverage guarantee is retained because the same factor is applied in calibration. Combined with
  PIWR we refer to the full pipeline as \emph{PIWR-RACC} (physics-informed weak regression with residual-adaptive
  conformal calibration).

We test these methods on all eleven DINN benchmarks with the original data-generating settings, without noise
and with $1\,\%$ noise, and compare against the DINN results as reported in \citet{Shaier2022DINN}. We extend the study with two cases from outside the DINN set: the seven-compartment S3I3R model of
\citet{Nielsen2025PIR}, with their PIR and PINN results as reference, and the real 1978 boarding-school influenza
outbreak \citep{BMJ1978Influenza}, with the Bayesian analysis in Stan of \citet{Grinsztajn2021Stan} as reference, and the time-varying transmission
benchmark of \citet{Millevoi2024PINN}, where $\beta(t)$ is expanded in B-splines and the results are compared with
physics-informed neural networks.

\section{Regression framework: parameter-linear epidemic models}\label{sec:models}

To present the regresion frameworks used in this study, the basis is that we consider models that an be described as a dynamical system
\begin{align}
\dot\bx = f(\bx, t; \bm\omega),
\end{align} 
where the the derivative with
respect to time is denoted as $\dot\bx(t) = \mathrm{d}\bx/\mathrm{d}t$, with $n$ states $\bx(t)\in\R^n$ and $k$ physical
parameters $\bm\omega\in\R^k$ that can be written as
\begin{equation}
\dot\bx(t) = A\big(\bx(t), t\big)\,\bm\theta, \qquad \bm\theta = g(\bm\omega) \in \R^p,
\label{eq:linear}
\end{equation}
where the library $A(\bx, t) \in \R^{n\times p}$ contains no unknowns. 

Together with an initial condition
$\bx(t_0) = \bx_0$, \eqref{eq:linear} is an initial value problem (IVP) on $t \in [t_0, T]$ with solution
$\bx(t; \bm\omega, \bx_0)$. We use it as the model of the data where the measured states are assumed to be this solution plus measurement noise, for some unknown $\bm\omega$ and $\bx_0$. For synthetic benchmarks where data is generated numerically this can hold by construction, however, for real data it is an assumption, and the fitted model reproduces the data only up to noise and model error. 

The initial state $\bx(t_0)$ is part of the model when stated as an IVP, but not of the physics-informed regression (PIR) \citet{Nielsen2025PIR} that uses only the derivatives and needs no $\bx_0$. The derivative may be approximated by finite differences, however, these are prone to noise when applied to real data and needs mitigation (fx. via filtering time series of the data) as explained in \citet{Nielsen2025PIR}. However,  
given the initial state, the general solution to the IVP may be found by temporal integration stated as
\begin{align}
\bx(t) = \bx(t_0) + \int_{t_0}^{t} f(\bx, t; \bm\omega) dt.
\end{align}
This formulation motivates a new {\em physics-informed weak regression (PIWR)} method that can estimate the intial condition $\bx_0$ jointly with $\bm\theta$ (\cref{sec:piwr}), so that the noise of the first sample does not enter every equation. Then the model may still be formulated as linear in the coefficient vector $\bm\theta$, while the map $g:\R^k\to\R^p$ from the physical parameters to the coefficients may be nonlinear. We assume a left inverse $h:\R^p\to\R^k$ with $h(g(\bm\omega)) = \bm\omega$ to be able to recover coefficients. This requires $p \ge k$ and for $p > k$ the coefficients satisfy $p - k$ constraints (\cref{sec:res:extra}). Known terms of the model enter as an offset, 
\begin{equation}
\dot\bx(t) = f_0(\bx, t) + A(\bx, t)\,\bm\theta,
\label{eq:linear_known}
\end{equation}
and are subtracted from the data side of the regression. In the
simplest case $\bm\theta = \bm\omega$. In many models \eqref{eq:linear_known} the right-hand side is nonlinear in $\bm\omega$ but linear in products or ratios of parameters, and then $\bm\theta$ collects these products and $\bm\omega = h(\bm\theta)$ is recovered in closed form. The structure also exposes {\em identifiability}: two parameters that enter \emph{only} as a product correspond to a single column of $A$ and cannot be separated by any method. One of them is then fixed. All eleven models, their coefficient vectors and the fixed parameters are given in \cref{app:models} and summarized in \cref{tab:models}.

The framework presented is general. Let's consider two worked examples to understand how to formulate  regression problems for parameter estimation.

\paragraph{Worked example: COVID-19 SIDR model.} The first DINN benchmark is the SIDR model with transmission
rate $\alpha$, recovery rate $\beta$, death rate $\gamma$ and known population $N = 59\cdot 10^6$,
\begin{equation}
\frac{d}{dt}\begin{bmatrix}S\\ I\\ D\\ R\end{bmatrix}
=\begin{bmatrix} -SI/N & 0 & 0\\ SI/N & -I & -I\\ 0 & 0 & I\\ 0 & I & 0 \end{bmatrix}
\begin{bmatrix}\alpha\\ \beta\\ \gamma\end{bmatrix},
\label{eq:sidr}
\end{equation}
i.e.\ $\dot S = -\alpha SI/N$, $\dot I = \alpha SI/N - (\beta+\gamma)I$, $\dot D = \gamma I$, $\dot R = \beta I$. The data
are 100 equally spaced samples on $[0, 500]$ days ($\Delta t = 5.05$ days) of the solution with
$(\alpha, \beta, \gamma) = (0.191, 0.05, 0.0294)$ from $S_0 = N - 1$, $I_0 = 1$. The outbreak rises and decays within
about 100 days, so it is covered by only about 20 samples. We use this example to explain every step below. $\square$

\paragraph{Worked example: a mapped coefficient vector (measles).} In the SIDR model every parameter multiplies its
own library term, so $\bm\theta = \bm\omega$. This is not the case in general. The seasonally forced measles model
of \citet{Shaier2022DINN} (\cref{app:measles}) has the state $\bx = (S, E, I)$, known population $N$, and the
parameters $\bm\omega = (\mu, \beta_0, \beta_1, \gamma, \sigma)$ (birth and death rate, mean transmission rate and its
seasonal amplitude, recovery rate, and a progression parameter), with the transmission rate
$\beta(t) = \beta_0(1 + \beta_1\cos 2\pi t)$. The equations
$\dot S = \mu(N - S) - \beta(t)SI/N$, $\dot E = \beta(t)SI/N - \mu\sigma E$ and $\dot I = \sigma E - (\mu + \gamma)I$
contain the products $\beta_0\beta_1$ and $\mu\sigma$, so they are nonlinear in $\bm\omega$. Expanding
$\beta(t)SI/N = \beta_0\,SI/N + \beta_0\beta_1\cos(2\pi t)\,SI/N$ and treating each product as one coefficient gives
the parameter-linear form \eqref{eq:linear},
\begin{equation}
\frac{d}{dt}\begin{bmatrix}S\\ E\\ I\end{bmatrix}
=\begin{bmatrix}
N - S & -SI/N & -\cos(2\pi t)\,SI/N & 0 & 0 & 0\\
0 & SI/N & \cos(2\pi t)\,SI/N & -E & 0 & 0\\
-I & 0 & 0 & 0 & E & -I
\end{bmatrix}
\begin{bmatrix}\mu\\ \beta_0\\ \beta_0\beta_1\\ \mu\sigma\\ \sigma\\ \gamma\end{bmatrix},
\label{eq:measles:example}
\end{equation}
with a library that depends explicitly on time through the known or assumed forcing. The coefficient map and its left inverse
are
\begin{equation}
\begin{aligned}
\bm\theta &= g(\bm\omega) = (\mu,\ \beta_0,\ \beta_0\beta_1,\ \mu\sigma,\ \sigma,\ \gamma) \in \R^6,\\
\bm\omega &= h(\bm\theta) = (\vartheta_1,\ \vartheta_2,\ \vartheta_3/\vartheta_2,\ \vartheta_6,\ \vartheta_5),
\end{aligned}
\label{eq:measles:maps}
\end{equation}
where $\vartheta_c$ denotes the $c$-th entry of $\bm\theta$. The regression problem is then formualted to be  solved for the six coefficients, and the
five physical parameters follow from $h(\bm\theta)$. For example, the seasonal amplitude is the ratio $\beta_1 = \vartheta_3/\vartheta_2$.
Since $p = 6 > k = 5$, the coefficients satisfy one constraint, $\vartheta_4 = \vartheta_1\vartheta_5$, which the linear
regression does not impose, and $h$ simply does not use $\vartheta_4$. The uncertainty of $\hat{\bm\theta}$ is carried
through $h(\bm\theta)$ by the delta method \eqref{eq:delta} (\cref{sec:delta}). The example serves only to explain how a {\em mapped coefficient vector} is formulated. $\square$

In the following, the physics-informed regression methods is explained with the SIDR model, and the measles results are given in \cref{app:measles}.

\begin{table}[htb]\centering\scriptsize
\caption{The eleven DINN benchmark models: number of states $n$, estimated physical parameters $k$, length $p$
of the coefficient vector $\bm\theta$, parameters fixed because they are not identifiable or absent from the
model, number of samples $m$, time horizon $T$, sampling interval $\Delta t$, and the DINN table and training
time reported by \citet{Shaier2022DINN}.}
\label{tab:models}
\setlength{\tabcolsep}{3.5pt}
\begin{tabular}{l r r r l r r r r r}
\toprule
Model & $n$ & $k$ & $p$ & fixed or absent & $m$ & $T$ [d] & $\Delta t$ [d] & DINN tab. & DINN time\\
\midrule
COVID-19 (SIDR) & 4 & 3 & 3 & -- & 100 & 500 & 5.05 & 8 & 20 min \\
HIV & 3 & 8 & 9 & $\mu_T$, $N$ & 50 & 20 & 0.408 & 9 & 22 h \\
Smallpox & 8 & 8 & 10 & -- & 50 & 100 & 2.04 & 10 & 14 h \\
Tuberculosis (SLIT) & 4 & 8 & 8 & $c$ & 50 & 40 & 0.816 & 11 & 12 h \\
Pneumonia (SVCIR) & 5 & 14 & 15 & $k$, $\tau$ & 50 & 100 & 2.04 & 12 & 41 h \\
Ebola (SEIHFR) & 6 & 10 & 10 & $\theta_1$, $\delta_1$, $\delta_2$ & 100 & 100 & 1.01 & 13 & 33 h \\
Dengue & 7 & 10 & 11 & $\lambda_v$ & 50 & 200 & 4.08 & 14 & 58 h \\
Anthrax (SIAC) & 4 & 12 & 12 & -- & 100 & 4000 & 40.4 & 16 & 91 h \\
Polio & 6 & 8 & 8 & -- & 50 & 40 & 0.816 & 18 & 66 h \\
Measles (seasonal SEI) & 3 & 5 & 6 & -- & 50 & 3650 & 74.5 & 20 & 28 h \\
Zika & 9 & 13 & 16 & $a$, $m$ & 50 & 30 & 0.612 & 22 & 13 h \\
\bottomrule
\end{tabular}
\end{table}

\section{Physics-informed regression}\label{sec:pir}

In this section, we detail the physics-informed regression methods used in this study. We omit the details of related techniques such as PINN \citep{RAISSI2019686} and DINN \citep{Shaier2022DINN} and only refer to reported results obtained using such methods.

\subsection{Strong form: physics-informed regression (PIR)}
Given measured samples $\tilde\bx_i$ of the states $\bx(t_i)$, which may contain noise (\cref{sec:setup}), $i = 0, \dots, m-1$, on an equispaced grid, PIR \citep{Nielsen2025PIR} replaces
$\dot\bx(t_i)$ in \eqref{eq:linear} by a finite-difference estimate $\mathbf d_i$ and stacks the equations of all
interior samples into the overdetermined system
\begin{equation}
\mathbf A\,\bm\theta \approx \mathbf b, \qquad
\mathbf A = \begin{bmatrix} A(\tilde\bx_{i_1}, t_{i_1})\\ \vdots\\ A(\tilde\bx_{i_M}, t_{i_M})\end{bmatrix},\quad
\mathbf b = \begin{bmatrix}\mathbf d_{i_1}\\ \vdots\\ \mathbf d_{i_M}\end{bmatrix},\qquad
\hat{\bm\theta} = \arg\min_{\bm\theta}\lVert\mathbf A\bm\theta - \mathbf b\rVert_2^2 .
\label{eq:pir}
\end{equation}
\citet{Nielsen2025PIR} use the second-order central difference (PIR-FD2). However, we also use the fourth- and sixth-order central differences,
\begin{equation}
\begin{aligned}
\mathbf d_i^{(2)} &= \frac{\tilde\bx_{i+1}-\tilde\bx_{i-1}}{2\Delta t},\qquad
\mathbf d_i^{(4)} = \frac{\tilde\bx_{i-2}-8\tilde\bx_{i-1}+8\tilde\bx_{i+1}-\tilde\bx_{i+2}}{12\Delta t},\\
\mathbf d_i^{(6)} &= \frac{-\tilde\bx_{i-3}+9\tilde\bx_{i-2}-45\tilde\bx_{i-1}+45\tilde\bx_{i+1}-9\tilde\bx_{i+2}+\tilde\bx_{i+3}}{60\Delta t},
\end{aligned}
\label{eq:fd}
\end{equation}
with truncation errors $\mathcal O(\Delta t^2)$, $\mathcal O(\Delta t^4)$, $\mathcal O(\Delta t^6)$, at the interior points where
the stencil fits. For white noise of standard deviation $\sigma$ the derivative noise has standard deviation
$0.71\,\sigma/\Delta t$, $0.95\,\sigma/\Delta t$ and $1.08\,\sigma/\Delta t$, respectively: higher orders remove truncation bias
but amplify noise. Two numerical details matter. First, each state equation may be divided by the magnitude
of its compartment, so that compartments of different size weigh comparably. With $s_c = \max_i\lvert\tilde x_{i,c}\rvert$,
$c = 1, \dots, n$, this is the row scaling
\begin{equation}
\mathbf D_n = \mathrm{diag}\big(1/s_1, \dots, 1/s_n\big), \qquad
\mathbf A_D = (\mathbf I_M \otimes \mathbf D_n)\,\mathbf A, \qquad
\mathbf b_D = (\mathbf I_M \otimes \mathbf D_n)\,\mathbf b,
\label{eq:rowscale}
\end{equation}
where $M$ is the number of interior samples $i_1, \dots, i_M$ in \eqref{eq:pir} at which the stencil fits,
$\mathbf I_M \in \R^{M\times M}$ is the identity matrix and $\otimes$ is the Kronecker product, so that
$\mathbf I_M \otimes \mathbf D_n = \mathrm{diag}(\mathbf D_n, \dots, \mathbf D_n) \in \R^{nM\times nM}$ is block diagonal with $M$ copies
of $\mathbf D_n$ and scales the $n$ equations of every sampled time in the same way. Row
scaling turns \eqref{eq:pir} into a weighted least-squares problem and therefore changes its minimizer; whether to use
it ($\mathbf D_n = \mathbf I_n$ otherwise) is decided per model and estimator on a tuning set (\cref{sec:setup}). Second, the
columns are scaled to unit norm before the solve,
\begin{equation}
\mathbf C = \mathrm{diag}\big(1/\lVert(\mathbf A_D)_{:,1}\rVert_2, \dots, 1/\lVert(\mathbf A_D)_{:,p}\rVert_2\big), \qquad
\hat{\mathbf u} = \arg\min_{\mathbf u}\lVert \mathbf A_D\mathbf C\,\mathbf u - \mathbf b_D\rVert_2^2, \qquad
\hat{\bm\theta} = \mathbf C\hat{\mathbf u}.
\label{eq:colscale}
\end{equation}
This is the change of variables $\bm\theta = \mathbf C\mathbf u$ with a diagonal, invertible $\mathbf C$, so it leaves
the minimizer unchanged, but it removes the extreme scaling of columns (fx. such as $VT^2$ in the HIV model \eqref{eq:app:hiv}) and improves
the conditioning of the solve. The covariance transforms accordingly, $\Sigma_\theta = \mathbf C\,\Sigma_{\mathbf u}\,\mathbf C$.

For the SIDR model, $\mathbf A$ has $4(m-2) = 392$ rows (PIR-FD2) and three columns, and $\hat{\bm\theta}$ is obtained in less than a millisecond in the benchmarked setup (\cref{tab:covid}).

\subsection{Weak form: physics-informed weak regression (PIWR)}\label{sec:piwr}
Integrating \eqref{eq:linear} from $t_0$ to $t_j$ gives
\begin{equation}
\bx(t_j) = \bx_0 + \mathbf J_j\,\bm\theta, \qquad \mathbf J_j = \int_{t_0}^{t_j} A\big(\bx(s), s\big)\,ds \in \R^{n\times p},
\qquad j = 0,\dots,m-1 .
\label{eq:weak}
\end{equation}
The relation is still linear in $\bm\theta$, but it contains the data themselves instead of their derivatives. Here $\bx_0 = \bx(t_0)$ is the initial state; in the regression below, the measured samples $\tilde\bx_j$ replace $\bx(t_j)$ on the left and the states $\bx(s)$ inside the integral.
We now describe how the integrals are evaluated, how the linear system is assembled, and how it is solved.

\paragraph{Quadrature.} Write $A_i = A(\tilde\bx_i, t_i) \in \R^{n\times p}$ for the library evaluated at the samples. Every
entry $(A)_{rc}$, $r = 1,\dots,n$, $c = 1,\dots,p$, is interpolated over $t_0,\dots,t_{m-1}$ by a cubic spline (not-a-knot
end conditions), and the spline is integrated exactly from $t_0$ to $t_j$. The spline is linear in its data, so the
quadrature can be written with weights $w_{ji}$ that depend only on the time grid
\begin{equation}
\mathbf J_j \approx \sum_{i=0}^{m-1} w_{ji}\,A_i, \qquad j = 0,\dots,m-1, \qquad \mathbf J_0 = \mathbf 0 ,
\label{eq:quad}
\end{equation}
with an error of $\mathcal O(\Delta t^4)$ for smooth integrands. For comparison, the trapezoidal rule would use
$w_{ji} = \Delta t\,(1 - \tfrac12\delta_{i0} - \tfrac12\delta_{ij})$ for $0 \le i \le j$ and $w_{ji} = 0$ otherwise, with an error of
$\mathcal O(\Delta t^2)$. The weights are computed once per time grid, and evaluating \eqref{eq:quad} costs $\mathcal O(m^2np)$
operations, which is negligible.

\paragraph{The linear system.} The initial state $\bx_0$ is treated as $n$ additional unknowns, so that the noise of the
first sample is not propagated into every equation. With the unknown vector $\mathbf z = (\bm\theta, \bx_0) \in \R^{p+n}$, the
$m$ relations \eqref{eq:weak} are stacked into
\begin{equation}
\underbrace{\begin{bmatrix} \mathbf J_0 & \mathbf I_n\\ \mathbf J_1 & \mathbf I_n\\ \vdots & \vdots\\ \mathbf J_{m-1} & \mathbf I_n\end{bmatrix}}_{\mathbf M\ \in\ \R^{nm\times(p+n)}}
\begin{bmatrix}\bm\theta\\ \bx_0\end{bmatrix}
\approx
\underbrace{\begin{bmatrix}\tilde\bx_0\\ \tilde\bx_1\\ \vdots\\ \tilde\bx_{m-1}\end{bmatrix}}_{\mathbf y\ \in\ \R^{nm}} ,
\label{eq:piwr:system}
\end{equation}
where $\mathbf I_n\in \R^{n \times n}$ is the identity matrix. The right-hand side $\mathbf y$ contains the measured states, including the
first one, $\tilde\bx_0$, which differs from the unknown initial state $\bx_0$ by the measurement noise. For the SIDR model ($n = 4$, $p = 3$, $m = 100$), $\mathbf M$ has $400$ rows and $7$ columns. Optionally, each state
equation is scaled by the magnitude of its compartment, $s_c = \max_i\lvert \tilde x_{i,c}\rvert$. This is the diagonal row scaling
$\mathbf D = \mathbf I_m \otimes \mathbf D_n$, with $\mathbf M_D = \mathbf D\mathbf M$ and $\mathbf y_D = \mathbf D\mathbf y$, so
that compartments of very different size weigh comparably.

\paragraph{Closed-form least-squares solution.} PIWR minimizes the residual of \eqref{eq:piwr:system},
\begin{equation}
\begin{bmatrix}\hat{\bm\theta}\\ \hat\bx_0\end{bmatrix} 
= \arg\min_{\bm\theta,\bx_0}\sum_{j=0}^{m-1}\big\lVert \mathbf D_n\big(\bx_0 + \mathbf J_j\bm\theta - \tilde\bx_j\big)\big\rVert_2^2
= \arg\min_{\mathbf z}\ \lVert\mathbf M_D\mathbf z - \mathbf y_D\rVert_2^2. %
\label{eq:piwr}
\end{equation}
Its unique minimizer, when $\mathbf M_D$ has full column rank, is given by the normal
equations,
\begin{equation}
\hat{\mathbf z} = \begin{bmatrix}\hat{\bm\theta}\\ \hat\bx_0\end{bmatrix} = \big(\mathbf M_D^T\mathbf M_D\big)^{-1}\mathbf M_D^T\mathbf y_D .
\label{eq:piwr:normal}
\end{equation}
The block structure of $\mathbf M$ allows $\bx_0$ to be eliminated explicitly. For fixed $\bm\theta$ the minimizer over $\bx_0$ is
$\bx_0 = \bar\bx - \bar{\mathbf J}\bm\theta$, with the time averages $\bar\bx = \frac1m\sum_j\tilde\bx_j$ and $\bar{\mathbf J} = \frac1m\sum_j\mathbf J_j$.
Inserting this gives a $p\times p$ system in centered quantities,
\begin{equation}
\hat{\bm\theta} = \left(\sum_{j=0}^{m-1}\big(\mathbf J_j - \bar{\mathbf J}\big)^T\mathbf S\,\big(\mathbf J_j - \bar{\mathbf J}\big)\right)^{-1}
\sum_{j=0}^{m-1}\big(\mathbf J_j - \bar{\mathbf J}\big)^T\mathbf S\,\big(\tilde\bx_j - \bar\bx\big), \qquad
\hat\bx_0 = \bar\bx - \bar{\mathbf J}\hat{\bm\theta},
\label{eq:piwr:closed}
\end{equation}
with $\mathbf S = \mathbf D_n^2$. In this form PIWR regresses the deviations of the states from their mean on the
deviations of the integrated library from its mean. The intercept $\bx_0$ only absorbs the offset. The matrix in
\eqref{eq:piwr:closed} is the Schur complement of the $\bx_0$-block of $\mathbf M_D^T\mathbf M_D$, and it is nonsingular exactly
when the parameters are identifiable from the sampled trajectory. The covariance used for the conformal shape
(\cref{sec:conformal}) is the $\bm\theta$-block of $\bar\sigma^2(\mathbf M_D^T\mathbf M_D)^{-1}$,
\begin{equation}
\Sigma_\theta = \bar\sigma^2\left(\sum_{j=0}^{m-1}\big(\mathbf J_j - \bar{\mathbf J}\big)^T\mathbf S\,\big(\mathbf J_j - \bar{\mathbf J}\big)\right)^{-1},
\qquad \bar\sigma^2 = \frac{\lVert\mathbf y_D - \mathbf M_D\hat{\mathbf z}\rVert_2^2}{nm - p - n} .
\label{eq:piwr:cov}
\end{equation}
In implementation, we do not form the normal equations \eqref{eq:piwr:normal}, whose condition number is the square of that of
$\mathbf M_D$. Instead, the columns are first scaled to unit norm, $\mathbf C = \mathrm{diag}(1/\lVert(\mathbf M_D)_{:,k}\rVert_2)$, which leaves the
minimizer unchanged. Then \eqref{eq:piwr} is solved by an orthogonal factorization: with $\mathbf M_D\mathbf C = \mathbf Q\mathbf R$,
$\hat{\mathbf z} = \mathbf C\mathbf R^{-1}\mathbf Q^T\mathbf y_D$ (in Python we can use the SVD-based \texttt{numpy.linalg.lstsq}, which also handles rank
deficiency). The same procedure, with $\mathbf M$ and $\mathbf y$ replaced by $\mathbf A$ and $\mathbf b$ of \eqref{eq:pir} and no intercept,
solves the strong-form problem of PIR.

\paragraph{Properties.} Compared with PIR, PIWR (i) uses all samples, including the end points;
(ii) averages the noise in the integrals instead of amplifying it in a difference quotient; and (iii) needs to
resolve only the \emph{area} under a sharp outbreak, not its slope. Its error still depends on how well the
sampled library resolves the dynamics. It also inherits an errors-in-variables bias, because the library is evaluated
at noisy states.

\subsection{From coefficients to parameters}\label{sec:delta}
Both estimators return $\hat{\bm\theta}$ and the ordinary least-squares covariance
$\Sigma_\theta = \bar\sigma^2(\mathbf A^T\mathbf A)^{-1}$, $\bar\sigma^2 = \lVert\mathbf r\rVert^2/(\#\text{rows}-\#\text{unknowns})$, with $\mathbf r$
the residual (for PIWR, the $\bm\theta$-block of the covariance of $(\bm\theta, \bx_0)$). The physical parameters are
$\hat{\bm\omega} = h(\hat{\bm\theta})$. Their uncertainty follows from the {\em delta method}, i.e.\ first-order
error propagation through the map $h$. A first-order Taylor expansion about $\hat{\bm\theta}$ gives
$h(\bm\theta) \approx h(\hat{\bm\theta}) + J_h(\bm\theta - \hat{\bm\theta})$, so that the covariance of
$\hat{\bm\omega}$ is approximately $J_h\Sigma_\theta J_h^T$. Here $J_h \in \R^{k\times p}$ is the Jacobian matrix of
$h$ at the estimate, with entries $(J_h)_{ij} = \partial h_i/\partial\vartheta_j$, the sensitivity of parameter
$\omega_i$ to coefficient $\vartheta_j$; it is computed by central differences, and it is the identity when
$\bm\theta = \bm\omega$. The standard error of $\hat\omega_j$ is
\begin{equation}
\hat\sigma_j^2 = \big[J_h\,\Sigma_\theta\,J_h^T\big]_{jj}, \qquad (J_h)_{ij} = \frac{\partial h_i}{\partial\vartheta_j}\bigg|_{\bm\theta = \hat{\bm\theta}} .
\label{eq:delta}
\end{equation}
The $\hat\sigma_j$ treat the residual as white noise. They ignore the smooth truncation bias of the derivative or
quadrature and the correlation of the residuals, and are therefore not trustworthy as uncertainties. They are, however, a good \emph{shape} since they encode how well each parameter is determined relative to the others. The conformal layer of the next section turns them into intervals with guaranteed coverage.

\section{Conformal prediction with physics-based models}\label{sec:conformal}

\subsection{A calibration population from the model}
Split conformal prediction \citep{vovk2005algorithmic,lei2018distribution,angelopoulos2023gentle} requires
calibration examples with known truth that are exchangeable with the test case. In standard regression these
are held-out data points. A model coefficient has no such examples: each data set has
exactly one true parameter vector, and calibrating $\lvert\hat\omega_j - \omega_j\rvert$ requires a population of data
sets with different known parameters. For a physics-based model this population is cheap to generate, and
generating it is the central idea of our extension. Let $\mathcal D$ be a prior box for the parameters. For
$i = 1, \dots, n$:
\begin{enumerate}
\item draw $\bm\omega^{(i)} \sim U(\mathcal D)$;
\item simulate the model from the same initial condition on the same time grid as the data;
\item add noise with the same noise model as the data;
\item run the \emph{fixed} estimator (PIR or PIWR, with the same settings) to obtain $\hat{\bm\omega}^{(i)}$ and
  $\hat{\bm\sigma}^{(i)}$.
\end{enumerate}
The estimator is never refitted. All of its tuning (finite-difference order, row scaling, reference residual)
is done on separate tuning draws, so that the calibration scores stay exchangeable with the score of a new
problem from the same population.

\subsection{Simultaneous intervals}
To obtain intervals that hold for all $k$ parameters at once, we use the maximum normalized error as the
nonconformity score,
\begin{equation}
s^{(i)} = \max_{j=1,\dots,k}\ \frac{\lvert\omega^{(i)}_j - \hat\omega^{(i)}_j\rvert}{\hat\sigma^{(i)}_j\,h^{(i)}},
\label{eq:score}
\end{equation}
with $h^{(i)} = 1$ for the plain layer. With $\hat q$ the $\lceil(n+1)(1-\alpha)\rceil$-th smallest score, the intervals
for a new data set are
\begin{equation}
\mathcal C_j = \big[\hat\omega_j - \hat q\,\hat\sigma_j h,\ \hat\omega_j + \hat q\,\hat\sigma_j h\big], \qquad j = 1,\dots,k,
\label{eq:interval}
\end{equation}
and, if the new problem is exchangeable with the calibration problems,
\begin{equation}
\mathbb P\big(\omega_j \in \mathcal C_j \ \text{for all}\ j = 1,\dots,k\big) \ge 1 - \alpha .
\label{eq:coverage}
\end{equation}
The guarantee holds for any $n$, any noise distribution and any bias of the estimator; a poor shape costs width
but never coverage. Two properties are worth stating explicitly. First, the guarantee is \emph{marginal over the
population}: it holds on average over parameters drawn from $\mathcal D$, so a wide prior box yields intervals sized
for the hardest epidemics in the box. Second, it is a statement about the estimator \emph{and} the model: if the
data were generated by a different model, the intervals carry no guarantee.

\subsection{Residual-adaptive calibration and PIWR-RACC}
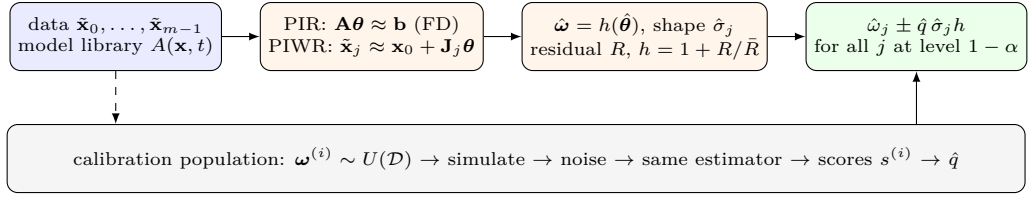
\begin{figure}[tb]\centering
\begin{tikzpicture}[node distance=4mm and 5mm, every node/.style={font=\scriptsize},
  box/.style={draw, rounded corners, align=center, minimum height=9mm, inner sep=3pt, fill=#1!8},
  box/.default=gray, >=latex]
\node[box=blue] (data) {data $\tilde\bx_0,\dots,\tilde\bx_{m-1}$\\ model library $A(\bx, t)$};
\node[box=orange, right=of data] (est) {PIR: $\mathbf A\bm\theta\approx\mathbf b$ (FD)\\ PIWR: $\tilde\bx_j \approx \bx_0 + \mathbf J_j\bm\theta$};
\node[box=orange, right=of est] (shape) {$\hat{\bm\omega} = h(\hat{\bm\theta})$, shape $\hat\sigma_j$\\ residual $R$, $h = 1 + R/\bar R$};
\node[box=green, right=of shape] (out) {$\hat\omega_j \pm \hat q\,\hat\sigma_j h$\\ for all $j$ at level $1-\alpha$};
\path let \p1 = (data.west), \p2 = (out.east) in
  node[box=gray, anchor=north, minimum width=\x2-\x1] (pop) at ($(data.south)!0.5!(out.south) + (0,-7mm)$)
  {calibration population: $\bm\omega^{(i)}\sim U(\mathcal D)$ $\rightarrow$
  simulate $\rightarrow$ noise $\rightarrow$ same estimator $\rightarrow$ scores $s^{(i)}$ $\rightarrow$ $\hat q$};
\draw[->] (data) -- (est); \draw[->] (est) -- (shape); \draw[->] (shape) -- (out);
\draw[->] (pop.north -| out.south) -- (out.south);
\draw[->, dashed] (data.south) -- (data.south |- pop.north);
\end{tikzpicture}
\caption{Conformalized physics-informed regression. The same fixed estimator (PIR or PIWR) is applied to the data and to
a population of simulated epidemics with known parameters; the calibration scores of the population set the scale
$\hat q$ of the simultaneous intervals.}
\label{fig:pipeline}
\end{figure}

The relative residual of the regression,
\begin{equation}
R = \lVert\mathbf b - \mathbf A\hat{\bm\theta}\rVert / \lVert\mathbf b\rVert ,
\end{equation}
measures how well the data satisfy the discretized equations. It grows when the sampling does not resolve the
dynamics or the noise is large, and it is available without the true parameters. We therefore introduce the
residual-adaptive factor
\begin{equation}
h = 1 + R/\bar R ,
\label{eq:h}
\end{equation}
with $\bar R$ the median residual on 200 tuning draws, and use it in the score \eqref{eq:score} and the intervals
\eqref{eq:interval}. Since $h$ depends only on the data and the fixed estimator, \eqref{eq:coverage} is unchanged.
Problems that the residual flags as difficult receive wider intervals and easy ones narrower intervals, which
moves the coverage from marginal towards conditional. Combined with PIWR we call the conformalized model PIWR-RACC. It is
summarized in \cref{fig:pipeline}.

\section{Results}\label{sec:results}

\subsection{Setup}\label{sec:setup}
For each of the eleven DINN benchmarks we regenerate the data with the original initial conditions, time grids and
parameter values (\cref{tab:models}, \cref{app:models}), and add noise per compartment $c$ as
$\tilde x_c(t_i) = x_c(t_i) + \epsilon\max_t\lvert x_c(t)\rvert\,\xi_i$, $\xi_i \sim \mathcal N(0,1)$ \citep{Nielsen2025PIR}, with
$\epsilon = 0$ (the setting of \citet{Shaier2022DINN}) and $\epsilon = 1\,\%$. We compare PIR-FD2, PIR-FD4, PIR-FD6 and
PIWR, each with plain and residual-adaptive calibration. The prior box is $\pm 50\,\%$ around the literature
values (capped at 1 for probabilities, and $[0, c]$ for the three parameters whose literature value is zero),
which is much wider than the DINN search ranges. Every estimator is tuned on draws that are not used for
calibration or testing: the row scaling on 100 draws, $\bar R$ on 200 draws. The calibration uses $n = 999$
problems and $\alpha = 0.1$. The coverage is measured on 2,000 independent test problems from the same population,
and, for $\epsilon = 1\,\%$, on 1,000 noise realizations at the literature values. Errors are reported in percent of
the actual value, as the median over the estimated parameters of a model.

\subsection{Worked example: COVID-19}\label{sec:res:covid}
\begin{figure}[tb]\centering
\includegraphics[width=\textwidth]{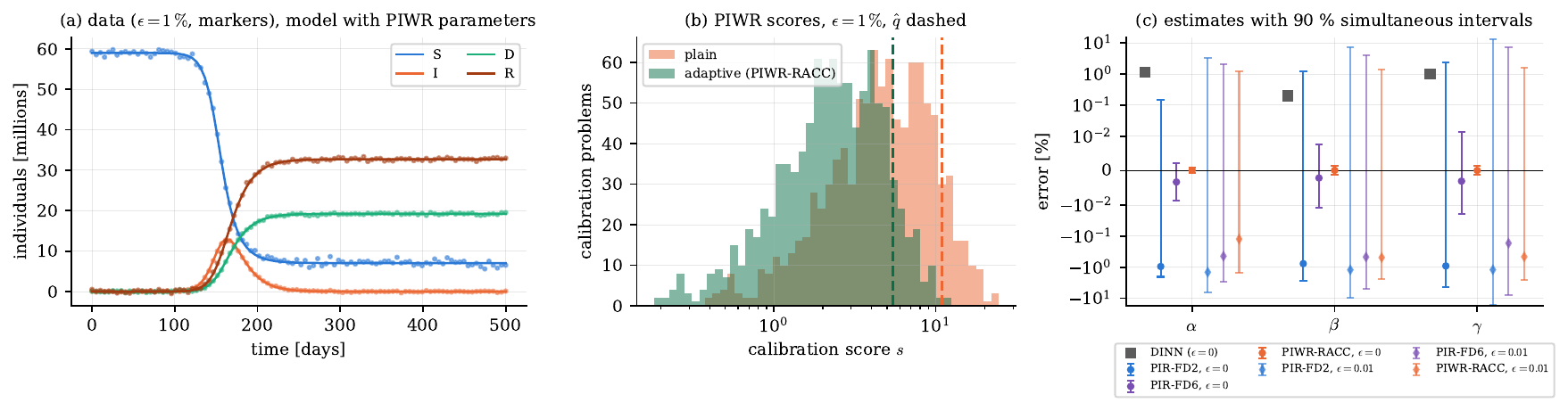}
\caption{COVID-19 (SIDR). (a) Data with $1\,\%$ noise (markers) and the model integrated with the PIWR parameters
(lines). (b) Calibration scores of PIWR with $1\,\%$ noise, plain and adaptive, with the quantiles $\hat q$ (dashed).
(c) Errors of the estimates with their 90\,\% simultaneous intervals (plain for PIR, PIWR-RACC for PIWR) for both
noise levels, next to the DINN errors; symmetric logarithmic axis.}
\label{fig:covid}
\end{figure}

\Cref{tab:covid} and \cref{fig:covid} follow the procedure of \cref{sec:pir,sec:conformal} for the SIDR model.
Without noise, the error of PIR-FD2, $0.76$--$0.95\,\%$, is the truncation error of the central difference at
$\Delta t = 5.05$ days. It is comparable to the DINN errors ($0.2$--$1.15\,\%$). PIR-FD6 reduces it to $0.002$--$0.003\,\%$,
and PIWR to $0.0001\,\%$, four orders of magnitude below DINN, from a least-squares solve of less than a millisecond (\cref{tab:covid}).
The PIWR-RACC intervals are correspondingly narrow (median half-width $\pm 0.001\,\%$) and contain the literature values.
With $1\,\%$ noise the error of PIR-FD2 grows to $1.2$--$1.5\,\%$, while PIR-FD6 and PIWR reach $0.12$--$0.49\,\%$. The
intervals of PIWR are about half as wide as those of PIR-FD6 (median $\pm 2.1\,\%$ against $\pm 4.5\,\%$, and $\pm 1.9\,\%$ with
PIWR-RACC), because integration averages the noise instead of amplifying it. \Cref{fig:covid}(b) shows the calibration scores: dividing by $h$ shifts the scores of the poorly resolved draws, and the adaptive quantile $\hat q$ is about half the plain one indicating reduced uncertainty for the well-resolved estimates.

\begin{table}[tb]\centering\scriptsize
\caption{COVID-19 (SIDR): estimates found by PIR-FD2, PIR-FD4, PIR-FD6 and PIWR with their \% errors, and the
PIWR-RACC 90\,\% simultaneous interval, without noise and with $1\,\%$ noise. The CPU time is the median of 30
repeated fits of one data set (\cref{sec:res:cost}). DINN values and training time from \citet{Shaier2022DINN}
(noise-free data only).}
\label{tab:covid}
\setlength{\tabcolsep}{3pt}
\begin{tabular}{l r r r r r r r r r r r c}
\toprule
 & & \multicolumn{2}{c}{DINN} & \multicolumn{2}{c}{PIR-FD2} & \multicolumn{2}{c}{PIR-FD4} & \multicolumn{2}{c}{PIR-FD6}
 & \multicolumn{3}{c}{PIWR / PIWR-RACC}\\
\cmidrule(lr){3-4}\cmidrule(lr){5-6}\cmidrule(lr){7-8}\cmidrule(lr){9-10}\cmidrule(lr){11-13}
Par. & Actual & Found & \% err. & Found & \% err. & Found & \% err. & Found & \% err. & Found & \% err. & 90\,\% interval\\
\midrule
\multicolumn{13}{l}{\emph{noise level $\epsilon = 0$}} \\
$\alpha$ & 0.191 & 0.1932 & 1.15 & 0.1892 & 0.946 & 0.1909 & 0.0389 & 0.191 & 0.00329 & 0.191 & 0.000105 & [0.191, 0.191] \\
$\beta$ & 0.05 & 0.0501 & 0.2 & 0.04962 & 0.764 & 0.04999 & 0.0274 & 0.05 & 0.00212 & 0.05 & 0.000108 & [0.049999, 0.050001] \\
$\gamma$ & 0.0294 & 0.0297 & 1.02 & 0.02913 & 0.907 & 0.02939 & 0.0365 & 0.0294 & 0.00304 & 0.0294 & 7.05e-05 & [0.0294, 0.0294] \\
\multicolumn{2}{l}{CPU time [ms]} & \multicolumn{2}{c}{$\approx$20\,min} & \multicolumn{2}{c}{0.07} & \multicolumn{2}{c}{0.07} & \multicolumn{2}{c}{0.07} & \multicolumn{2}{c}{0.21} &  \\
\multicolumn{13}{l}{\emph{noise level $\epsilon = 0.01$}} \\
$\alpha$ & 0.191 &  &  & 0.1882 & 1.45 & 0.19 & 0.499 & 0.1902 & 0.441 & 0.1908 & 0.124 & [0.18817, 0.19335] \\
$\beta$ & 0.05 &  &  & 0.04939 & 1.21 & 0.04975 & 0.492 & 0.04977 & 0.468 & 0.04976 & 0.487 & [0.048807, 0.050706] \\
$\gamma$ & 0.0294 &  &  & 0.02905 & 1.2 & 0.02932 & 0.266 & 0.02935 & 0.168 & 0.02926 & 0.46 & [0.028645, 0.029884] \\
\multicolumn{2}{l}{CPU time [ms]} & \multicolumn{2}{c}{} & \multicolumn{2}{c}{0.07} & \multicolumn{2}{c}{0.07} & \multicolumn{2}{c}{0.07} & \multicolumn{2}{c}{0.22} &  \\
\bottomrule
\end{tabular}
\end{table}

\subsection{Accuracy against the sampling interval}\label{sec:res:conv}

\begin{figure}[tb]\centering
\includegraphics[width=\textwidth]{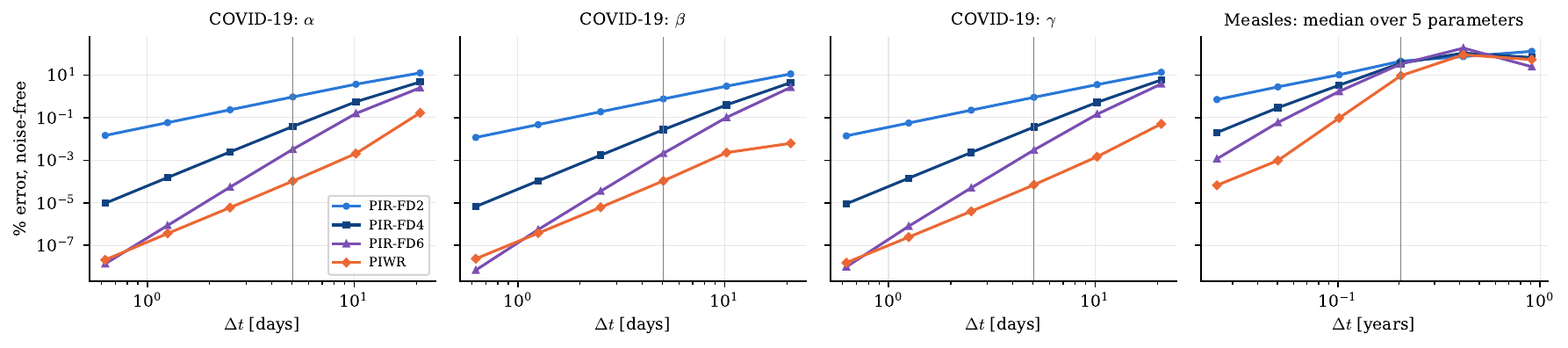}
\caption{Noise-free error against the sampling interval $\Delta t$ on the same time horizon. Results of the COVID-19 case (left 3 sub-figures) with error of each
parameter. Results of the Measles case with median error over its five parameters. The vertical line marks the DINN grid.}
\label{fig:conv}
\end{figure}

Since the finite difference approximation of the rate of change term in \eqref{eq:linear} introduced approximatinos that depend on the time grid we investigate the effect on two Models. \Cref{fig:conv} refines the time grid of the noise-free COVID-19 and Measles data. For COVID-19 the finite-difference estimators follow their design orders 2, 4 and 6 down to $10^{-8}\,\%$. PIWR converges at fourth order but with a much smaller constant, i.e. at the DINN grid it is 30 times more accurate than PIR-FD6 and $10^4$ times more accurate than PIR-FD2. PIR-FD6 overtakes it only on grids eight times finer. Measles is the extreme case: the outbreak is over within about $0.3$ years, and at the DINN spacing $\Delta t = 0.2$ years no finite-difference estimator is better than $33\,\%$. PIWR gives $9.4\,\%$ at the same spacing, because it needs to resolve only the area under the outbreak. It reaches $0.1\,\%$ when $\Delta t$ is halved and $10^{-3}\,\%$ when it is halved twice. PIR-FD6 needs $\Delta t \approx 0.025$ years for $10^{-3}\,\%$.

\subsection{Bechmarks across all eleven models}\label{sec:res:all}
\begin{figure}[tb]\centering
\includegraphics[width=\textwidth]{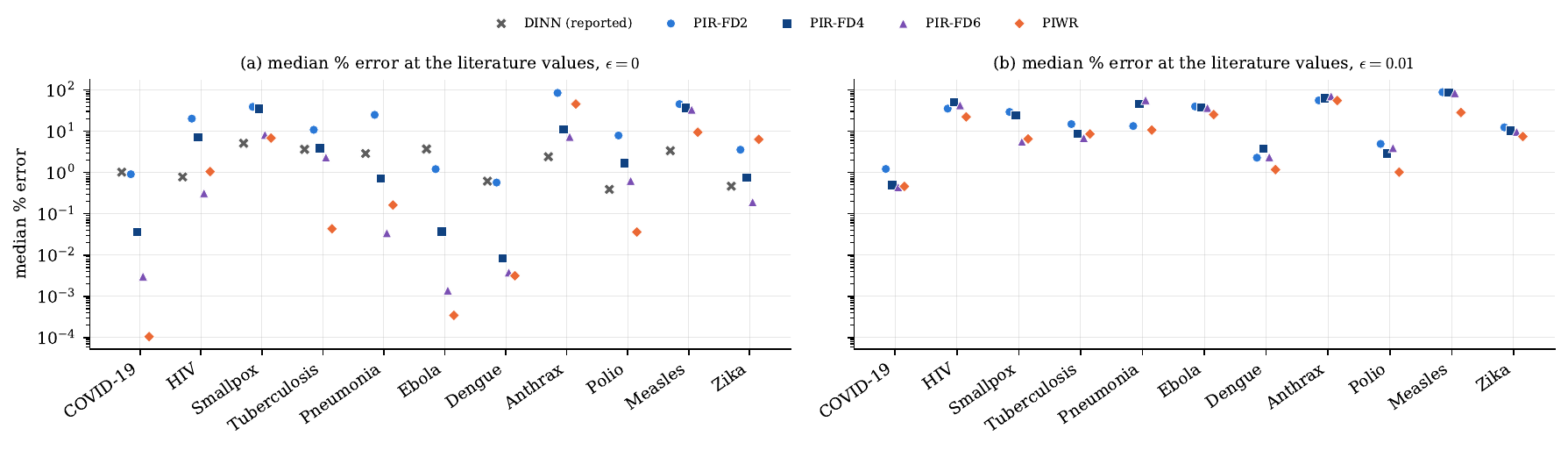}
\caption{Median \% error at the literature values for all models, without noise (a) and with $1\,\%$ noise (b).}
\label{fig:accuracy}
\end{figure}

We compare the results of the physics-informed regression models in 
\Cref{tab:accuracy} and \cref{fig:accuracy} summarize the accuracy for all models. We find that PIWR is the most accurate estimator in 15 of the 22 model/noise settings, and PIR-FD6 in the remaining seven. Without noise, PIWR or PIR-FD6 improves on the reported DINN errors by one to four orders of magnitude for COVID-19, Ebola, Dengue, Pneumonia, Polio and Tuberculosis, is comparable for HIV and Zika, and less accurate for Smallpox ($6.9\,\%$ against $5.1\,\%$) and Measles ($9.4\,\%$ against $3.4\,\%$). 

With $1\,\%$ noise the
advantage of PIWR is largest where the finite differences amplify the noise of small compartments: Pneumonia ($11\,\%$ against $13$--$57\,\%$), Ebola ($25\,\%$ against $37$--$40\,\%$), Measles ($28\,\%$ against $84$--$88\,\%$) and HIV ($22\,\%$ against $35$--$51\,\%$).

Two benchmarks remain out of reach at their DINN sampling. Anthrax has outbreaks of a few days, sampled every 40
days, and all estimators have median errors of $7$--$85\,\%$. For Zika, the row scaling selected on the tuning
draws for PIWR is suboptimal for the literature case ($6.3\,\%$ against $0.19\,\%$ for PIR-FD6 without noise).

The DINN errors are not directly comparable. \citet{Shaier2022DINN} restrict every parameter to a search range that
is typically $\pm 1$--$10\,\%$ around the literature value, which bounds their errors by construction. They also report
estimates for parameters that do not occur in the equations ($k$, $\tau$, $\lambda_v$, $m$) or that cannot be separated
from others ($\mu_T$ and $N$, $c$, $\theta_1$, $\delta_1$, $\delta_2$, $a$; \cref{tab:models}). PIR and PIWR search an unbounded
parameter space, and the least-squares structure makes identifiability explicit as collinear columns of the library.

\begin{table}[tb]\centering\scriptsize
\caption{Median \% error at the literature values over the estimated parameters of each model. DINN: noise-free
data, as reported by \citet{Shaier2022DINN} within prescribed search ranges. Bold: most accurate PIR/PIWR
estimator per noise level.}
\label{tab:accuracy}
\begin{tabular}{l r r r r r r r r r}
\toprule
 & & \multicolumn{4}{c}{$\epsilon = 0$} & \multicolumn{4}{c}{$\epsilon = 1\,\%$}\\
\cmidrule(lr){3-6}\cmidrule(lr){7-10}
Model & DINN & PIR-FD2 & PIR-FD4 & PIR-FD6 & PIWR & PIR-FD2 & PIR-FD4 & PIR-FD6 & PIWR\\
\midrule
COVID-19 & 1.02 & 0.91 & 0.036 & 0.003 & \textbf{0.00011} & 1.2 & 0.49 & \textbf{0.44} & 0.46 \\
HIV & 0.779 & 20 & 7.1 & \textbf{0.32} & 1.1 & 35 & 51 & 43 & \textbf{22} \\
Smallpox & 5.1 & 39 & 35 & 8.2 & \textbf{6.9} & 29 & 24 & \textbf{5.6} & 6.5 \\
Tuberculosis & 3.63 & 11 & 3.9 & 2.3 & \textbf{0.043} & 15 & 8.6 & \textbf{6.9} & 8.5 \\
Pneumonia & 2.88 & 25 & 0.72 & \textbf{0.034} & 0.16 & 13 & 45 & 57 & \textbf{11} \\
Ebola & 3.7 & 1.2 & 0.037 & 0.0014 & \textbf{0.00035} & 40 & 38 & 37 & \textbf{25} \\
Dengue & 0.616 & 0.57 & 0.0085 & 0.0038 & \textbf{0.0031} & 2.3 & 3.8 & 2.4 & \textbf{1.2} \\
Anthrax & 2.4 & 85 & 11 & \textbf{7.2} & 45 & 56 & 62 & 70 & \textbf{55} \\
Polio & 0.388 & 7.9 & 1.7 & 0.63 & \textbf{0.036} & 4.9 & 2.9 & 3.9 & \textbf{1} \\
Measles & 3.36 & 45 & 37 & 33 & \textbf{9.4} & 88 & 85 & 84 & \textbf{28} \\
Zika & 0.465 & 3.6 & 0.75 & \textbf{0.19} & 6.3 & 12 & 10 & 9.7 & \textbf{7.5} \\
\bottomrule
\end{tabular}
\end{table}

\subsection{Uncertainty quantification}\label{sec:res:uq}
\begin{figure}[tb]\centering
\includegraphics[width=\textwidth]{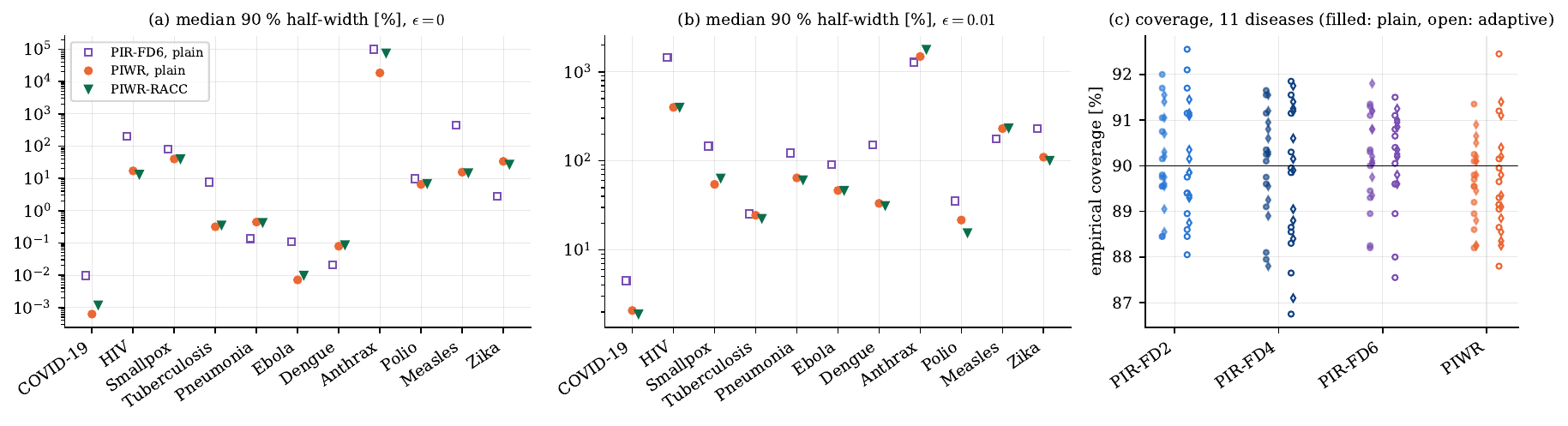}
\caption{(a, b) Median half-width of the 90\,\% simultaneous intervals at the literature values for PIR-FD6
(plain), PIWR (plain) and PIWR-RACC. (c) Empirical coverage on 2,000 test problems for all models and noise levels,
plain (filled) and adaptive (open); circles $\epsilon = 0$, diamonds $\epsilon = 1\,\%$. Target: 90\,\%.}
\label{fig:uq}
\end{figure}

While the DINN benchmarks report point estimates without uncertainty estimates, we equip every estimate with simultaneous intervals from the conformal prediction framework of \cref{sec:conformal}, calibrated on a population of
simulated epidemics. The uncertainty is modelled by simulation: each calibration epidemic draws its parameters
from the $\pm 50\,\%$ box around the literature values, is sampled on the same time grid, receives the measurement
noise of \cref{sec:setup}, and is passed through the same estimator. The calibrated intervals therefore cover every
error that this pipeline produces, from measurement noise as well as from the finite-difference truncation or the
quadrature of the weak form, and a $90\,\%$ interval means that, for a new data set drawn from the same population,
all parameters lie inside their intervals simultaneously with probability at least $90\,\%$. This is a frequentist
guarantee over epidemics and noise realizations, not a Bayesian posterior for one data set, and it assumes that the
compartment model itself is correct. Simulating this population ($1{,}199$ tuning and calibration epidemics per
model and noise level) is the dominant cost of the pipeline, seconds to minutes instead of the millisecond of a single
fit (\cref{sec:res:cost}), but it is what makes the intervals reliable. Their coverage is calibrated and can be
checked on independent test problems instead of resting on asymptotic assumptions. \Cref{tab:uq} and \cref{fig:uq} report the half-widths of the 90\,\% simultaneous intervals at the literature values
and the coverage on the test problems.

\textbf{Coverage.} Over all 88 settings (11 models, 4 estimators, 2 noise levels) the empirical coverage is
  $87.6$--$92.0\,\%$ for the plain layer (mean $90.0\,\%$) and $86.7$--$92.5\,\%$ for the adaptive layer (mean $89.9\,\%$).
  This is the target within the fluctuation expected for a single calibration set of 999 problems.

\textbf{Width follows accuracy.} The intervals are narrow where the estimator is accurate: for example, with PIWR-RACC
  and without noise, COVID-19 $\pm 0.001\,\%$, Ebola $\pm 0.01\,\%$, Dengue $\pm 0.09\,\%$, Tuberculosis $\pm 0.4\,\%$.
  They are wide or vacuous where the data do not resolve the dynamics (Anthrax, HIV with noise). No information about
  the true parameters is needed to tell these cases apart.

\textbf{PIWR narrows the intervals with noise.} With $1\,\%$ noise the PIWR intervals are narrower than those of
  PIR-FD6 for nine of the eleven models, e.g.\ Dengue $\pm 33\,\%$ against $\pm 150\,\%$, Smallpox $\pm 54\,\%$ against $\pm 147\,\%$,
  Ebola $\pm 46\,\%$ against $\pm 91\,\%$.

\textbf{The adaptive factor helps where the population is heterogeneous.} It narrows the intervals where the
  calibration box contains much harder epidemics than the literature case and the residual separates them, e.g.\ Polio
  with noise $\pm 22 \rightarrow \pm 15\,\%$, HIV without noise $\pm 24 \rightarrow \pm 18\,\%$, Zika $\pm 34 \rightarrow \pm 27\,\%$. It
  widens them slightly where the literature case is already typical (COVID-19 and Ebola without noise). The
  least-squares shape already scales with the absolute residual, so $h$ is a second-order correction.

\textbf{Coverage at the literature values.} The guarantee is marginal over the prior box; the coverage of one fixed
  parameter vector over noise realizations is a conditional quantity that it does not control. For PIR-FD6 this coverage is
  96--100\,\% for all models but one, because the literature case is easier than the average epidemic in the box. For
  PIWR it is 86--100\,\% and the PIWR intervals are tighter and, for COVID-19, Pneumonia, Dengue and Measles, the
  literature case is slightly harder than the average. Conditional guarantees would require a narrower, problem-specific
  calibration population, or localized conformal methods.

\begin{table}[tb]\centering\scriptsize
\caption{Median half-width (in \% of the actual value) of the 90\,\% simultaneous intervals at the literature values
for PIR-FD6 (plain), PIWR (plain) and PIWR-RACC, the empirical PIWR-RACC coverage on 2,000 test problems, and the total time of
the conformal pipeline per model. The pipeline covers four estimators, both noise levels, the tuning, 999
calibration and 2,000 test problems, and 1,000 noise realizations.}
\label{tab:uq}
\begin{tabular}{l r r r r r r r r r}
\toprule
 & \multicolumn{4}{c}{$\epsilon = 0$} & \multicolumn{4}{c}{$\epsilon = 1\,\%$} & \\
\cmidrule(lr){2-5}\cmidrule(lr){6-9}
Model & PIR-FD6 & PIWR & PIWR-RACC & cov. [\%] & PIR-FD6 & PIWR & PIWR-RACC & cov. [\%] & time [s]\\
\midrule
COVID-19 & 0.0097 & 0.00062 & 0.0012 & 89.0 & 4.5 & 2.1 & 1.9 & 89.1 & 9 \\
HIV & 2e+02 & 17 & 13 & 89.1 & 1.5e+03 & 4e+02 & 4e+02 & 88.8 & 10 \\
Smallpox & 80 & 40 & 40 & 89.6 & 1.5e+02 & 54 & 64 & 90.4 & 19 \\
Tuberculosis & 7.7 & 0.32 & 0.36 & 88.6 & 25 & 24 & 23 & 90.2 & 15 \\
Pneumonia & 0.14 & 0.44 & 0.43 & 90.0 & 1.2e+02 & 64 & 61 & 88.2 & 14 \\
Ebola & 0.11 & 0.0071 & 0.01 & 92.5 & 91 & 46 & 47 & 89.8 & 18 \\
Dengue & 0.021 & 0.079 & 0.088 & 91.2 & 1.5e+02 & 33 & 32 & 88.3 & 16 \\
Anthrax & 1.8e+13 & 1.9e+04 & 7.7e+04 & 89.3 & 1.3e+03 & 1.5e+03 & 1.8e+03 & 89.3 & 41 \\
Polio & 9.8 & 6.5 & 6.9 & 90.1 & 35 & 22 & 16 & 91.4 & 56 \\
Measles & 4.4e+02 & 16 & 15 & 89.1 & 1.8e+02 & 2.3e+02 & 2.3e+02 & 88.5 & 36 \\
Zika & 2.8 & 34 & 27 & 87.8 & 2.3e+02 & 1.1e+02 & 1e+02 & 91.1 & 26 \\
\bottomrule
\end{tabular}
\end{table}

\subsection{Further benchmarks: S3I3R and real influenza data}\label{sec:res:extra}
Two cases from outside the DINN set test the methods on a larger compartment model with a published PIR and PINN
reference, and on real data with a published Bayesian reference. The models are given in \cref{app:extra}.

\paragraph{S3I3R.} The seven-compartment model of \citet{Nielsen2025PIR} splits the infected into three stages of
increasing severity (\cref{app:s3i3r}). We use their parameters and initial state, daily samples over the first 28
days, and two settings: the published one, in which the recovery rates $\gamma_1,\gamma_2,\gamma_3$ are known and
$(\beta,\phi_1,\phi_2,\theta)$ are estimated, and the harder one with all seven rates unknown. Known terms of the model
enter the regression as an offset, $\dot\bx = f_0(\bx,t) + A(\bx,t)\bm\theta$, which is subtracted from
the derivative estimates $\mathbf d_i$ in \eqref{eq:pir} (strong form) or, integrated, from $\tilde\bx_j$ in \eqref{eq:weak} (weak form). All other
settings are as in \cref{sec:setup}. \Cref{tab:s3i3r} shows the results.

The noise-free ($\epsilon=0$) PIR-FD2 reproduces the published PIR errors to within a factor of 1.3--10
  ($0.06$--$0.22\,\%$ against $0.02$--$0.11\,\%$. The published data were generated with a fixed-step Runge--Kutta method). PIR-FD6 and PIWR reduce the errors to $10^{-6}$--$3\cdot 10^{-5}\,\%$. That is about three orders of magnitude below PIR-FD2 and five ordes of magnitude below the PINN of \citet{Nielsen2025PIR} ($0.25$--$9.5\,\%$).
  
With $1\,\%$ noise and $\gamma_i$ known it is found that PIWR has the smallest median error, $0.51\,\%$ against $0.79$--$1.0\,\%$ for  PIR-FD2/4/6 (individual errors $0.07$--$0.77\,\%$). The PIWR-RACC intervals are $\pm 0.7$--$2.8\,\%$, half as wide as those of PIR-FD6, at
  $91\,\%$ coverage.

With $1\,\%$ noise and all rates unknown, it si found that during the first four weeks where the epidemic is still growing, $I_2$ and $I_3$ are small and dominated by the noise. Hence, $\gamma_2$ and $\gamma_3$ are then practically not identifiable (PIWR errors $93\,\%$ and $27\,\%$). The intervals show this rather than hide it, since they contain
  zero for $\gamma_2$ and $\gamma_3$ ($\pm 141\,\%$ and $\pm 170\,\%$). They stay at $\pm 1.6\,\%$ for $\beta$ and $\pm 2.2\,\%$ for $\theta$, and the coverage is $89\,\%$.

\begin{table}[tb]\centering\scriptsize
\caption{S3I3R model \citep{Nielsen2025PIR}: \% errors of the estimates and PIWR-RACC 90\,\% simultaneous intervals
($\epsilon = 1\,\%$). PIR and PINN: errors reported in Tables~1 and~4 of \citet{Nielsen2025PIR} for noise-free data
with $\gamma_{1,2,3}$ known.}
\label{tab:s3i3r}
\setlength{\tabcolsep}{4pt}
\begin{tabular}{l r r r r r r r r c}
\toprule
 & & \multicolumn{2}{c}{\citet{Nielsen2025PIR}} & \multicolumn{3}{c}{$\epsilon = 0$} & \multicolumn{3}{c}{$\epsilon = 1\,\%$}\\
\cmidrule(lr){3-4}\cmidrule(lr){5-7}\cmidrule(lr){8-10}
Par. & Actual & PIR & PINN & PIR-FD2 & PIR-FD6 & PIWR & PIR-FD6 & PIWR & PIWR-RACC 90\,\% interval\\
\midrule
\multicolumn{10}{l}{\emph{$\gamma_1, \gamma_2, \gamma_3$ known (setting of Nielsen et al.)}} \\
$\beta$ & 0.5 & 0.0378 & 0.248 & 0.0579 & 3.36e-06 & 1.76e-06 & 0.0562 & 0.0674 & [0.497, 0.504] \\
$\phi_1$ & 0.05 & 0.111 & 1.86 & 0.149 & 1.4e-06 & 1.07e-05 & 0.595 & 0.444 & [0.049, 0.0506] \\
$\phi_2$ & 0.05 & 0.0236 & 3.24 & 0.173 & 5.45e-06 & 1.83e-05 & 1.59 & 0.565 & [0.0489, 0.0505] \\
$\theta$ & 0.1 & 0.0222 & 9.51 & 0.224 & 1.16e-05 & 3.19e-05 & 0.978 & 0.767 & [0.0979, 0.104] \\
\multicolumn{10}{l}{\emph{all seven rates unknown}} \\
$\beta$ & 0.5 & -- & -- & 0.203 & 2.52e-06 & 1.42e-05 & 1.78 & 0.417 & [0.49, 0.506] \\
$\gamma_1$ & 0.333 & -- & -- & 0.202 & 8.35e-06 & 4.6e-06 & 2.21 & 2.23 & [0.324, 0.358] \\
$\gamma_2$ & 0.05 & -- & -- & 0.629 & 0.00028 & 0.000376 & 154 & 92.6 & [-0.0668, 0.0743] \\
$\gamma_3$ & 0.05 & -- & -- & 1.15 & 6.93e-06 & 0.000215 & 54.6 & 27.4 & [-0.0215, 0.149] \\
$\phi_1$ & 0.05 & -- & -- & 0.245 & 6.27e-05 & 0.000104 & 32.4 & 20.1 & [0.0245, 0.0554] \\
$\phi_2$ & 0.05 & -- & -- & 0.0382 & 6.68e-06 & 5.64e-05 & 8.21 & 4.26 & [0.0371, 0.0671] \\
$\theta$ & 0.1 & -- & -- & 0.225 & 1.15e-05 & 3.18e-05 & 0.981 & 0.765 & [0.0986, 0.103] \\
\bottomrule
\end{tabular}
\end{table}

\paragraph{Influenza in a boarding school.} In January 1978, 512 of the 763 boys of a British boarding school fell ill
with influenza \citep{BMJ1978Influenza}. The data are the number of boys in bed on each of 14 days. They are a
standard test case for Bayesian inference in compartment models, and we compare with the SIR fit of
\citet{Grinsztajn2021Stan} in Stan: negative-binomial observations, weakly informative priors,
$S(0) = 762$, $I(0) = 1$. Only $I$ is observed. Eliminating $S = N - I - R$ and $R = \gamma\int_0^t I\,ds$ gives
\begin{equation}\label{eq:school}
\dot I = (\beta - \gamma)\, I - \frac{\beta}{N}\, I^2 - \frac{\beta\gamma}{N}\, I\, C(t), \qquad C(t) = \int_0^t I\,ds,
\end{equation}
which is linear in $\bm\theta = (\beta-\gamma,\ \beta/N,\ \beta\gamma/N)$. $C$ is computed from the data with the same
cubic-spline quadrature as the weak form. Three coefficients carry two parameters. A free fit of $\bm\theta$ ignores
the constraint $\theta_3 = \theta_2(N\theta_2 - \theta_1)$, which is costly with 15 noisy samples. We therefore also
use \emph{PIWR-VP}, which imposes the constraint by {\em variable projection} (VP) \citep{Golub1973VarPro}.
The integrated
\eqref{eq:school}, $I_j = I_0 + \beta\,(P_{1,j} - \gamma P_{2,j}) - \gamma C_j$ with $P_1 = \int I(N-I)/N$ and
$P_2 = \int IC/N$, is linear in $(\beta, I_0)$ for fixed $\gamma$. So $\gamma$ minimizes the profiled residual
in one dimension, and $(\beta, I_0)$ follow from a two-column least-squares solve. In short, VP keeps the cheap linear solve for the parameters that
enter linearly and leaves only a one-dimensional search for the one that does not. The covariance is the Gauss--Newton
covariance of $(\beta,\gamma,I_0)$, and $\mathcal R_0 = \beta/\gamma$ follows by the delta method \eqref{eq:delta}.

\begin{figure}[tb]\centering
\includegraphics[width=\textwidth]{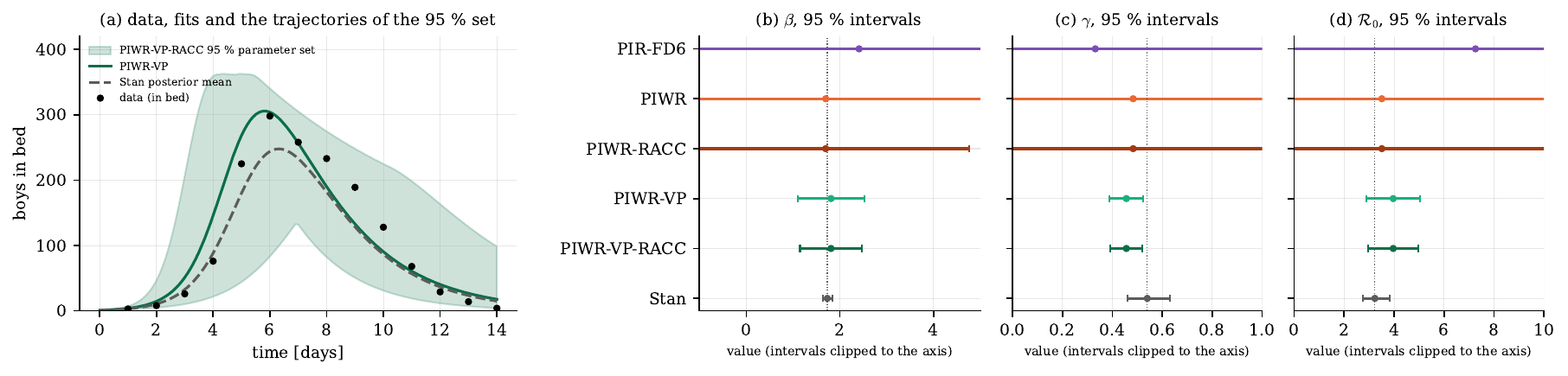}
\caption{Boarding-school influenza. (a) Number of boys in bed, the SIR trajectory with the PIWR-VP parameters, the
envelope of the trajectories whose parameters lie in the PIWR-VP-RACC 95\,\% set, and the trajectory at the Stan
posterior mean. (b--d) 95\,\% intervals for $\beta$, $\gamma$ and $\mathcal R_0$, where the dotted line is the Stan
posterior mean. The finite-difference and free-coefficient PIWR intervals extend beyond the axes.}
\label{fig:school}
\end{figure}

The calibration population is the SIR model with negative-binomial counts with mean $I$ and variance
$I + \phi^{-1}I^2$. $(\beta,\gamma,\phi^{-1})$ are drawn from a $\pm 50\,\%$ box around the pilot PIWR-VP fit to the
data, with $\phi^{-1} = 0.040$ from the moment estimate of the residuals. The intervals are simultaneous for
$(\beta,\gamma,\mathcal R_0)$ at $95\,\%$, to match the reported credible intervals. The coverage is measured on five
independent calibration/test splits. \Cref{tab:school} and \cref{fig:school} show the results. It is seen that 
imposing the coefficient constraint matters since with only 14 samples, fitting two parameters $(\beta, \gamma)$ rather than three free coefficients makes the problem better conditioned. The three coefficients satisfy
$\vartheta_3 = \vartheta_2(N\vartheta_2 - \vartheta_1)$ (\cref{app:school}), which the finite-difference and
free-coefficient PIWR estimates ignore. They are calibrated ($95$--$96\,\%$ coverage) but useless, since their intervals include negative rates. The median error on the test problems is $55\,\%$ for PIR-FD2 and $33\,\%$ for PIWR.
PIWR-VP reduces it to $7.6\,\%$, and its intervals are 5--12 times narrower: $\beta\in[1.15, 2.47]$,
$\gamma\in[0.39, 0.52]$ and $\mathcal R_0\in[2.97, 4.96]$ for PIWR-VP-RACC, at $94.4\,\%$ mean coverage
(93.3--96.6\,\% over the splits).

The PIWR-VP intervals contain the Stan posterior means of $\beta$ and $\mathcal R_0$, and the $\gamma$ interval
overlaps the credible interval. The estimate $\gamma = 0.46$ lies below the posterior mean $0.54$. An output-error
least-squares fit of the simulated SIR trajectory to the counts gives $\gamma = 0.44$ as well, so the difference comes
from the weighting of the observations (negative-binomial likelihood and priors in Stan, least squares here), not from
the weak form. The credible intervals are narrower, most of all for $\beta$, but they are conditional on the model,
the prior and the likelihood, and they are marginal for each parameter. The conformal intervals are simultaneous, and
their coverage is guaranteed over the calibration population rather than conditional on the data.

The cost remains small. A PIWR-VP fit takes 5\,ms, and the complete conformal pipeline, including five
calibration/test splits for five estimators, takes 110\,s.

\begin{table}[tb]\centering\scriptsize
\caption{Boarding-school influenza: estimates with 95\,\% simultaneous conformal intervals, mean coverage (range over
five calibration/test splits), median \% error on the test problems, and the time per fit, against the posterior
mean and 95\,\% credible intervals of the Stan SIR model \citep{Grinsztajn2021Stan}.}
\label{tab:school}
\setlength{\tabcolsep}{3pt}
\begin{tabular}{l c c c r r r}
\toprule
Method & $\beta$ [1/day] & $\gamma$ [1/day] & $\mathcal R_0$ & coverage [\%] & test err. [\%] & fit [ms]\\
\midrule
PIR-FD2 & 1.81 [-5.98, 9.61] & 0.4 [-1.14, 1.94] & 4.54 [-31.7, 40.8] & 95.6 (94.7--96.4) & 55.4 & 0.13 \\
PIR-FD6 & 2.41 [-7.15, 12] & 0.332 [-0.848, 1.51] & 7.26 [-45.9, 60.5] & 95.9 (94.8--96.9) & 57.8 & 0.13 \\
PIWR & 1.7 [-1.63, 5.02] & 0.483 [-0.356, 1.32] & 3.51 [-9.39, 16.4] & 95.7 (95.2--96.4) & 32.6 & 0.2 \\
PIWR-RACC & 1.7 [-1.36, 4.75] & 0.483 [-0.289, 1.26] & 3.51 [-8.36, 15.4] & 95.2 (94.3--95.5) & 32.6 & 0.2 \\
PIWR-VP & 1.81 [1.09, 2.52] & 0.456 [0.388, 0.524] & 3.96 [2.89, 5.04] & 94.4 (93.3--96.4) & 7.56 & 2.3 \\
PIWR-VP-RACC & 1.81 [1.15, 2.47] & 0.456 [0.393, 0.519] & 3.96 [2.97, 4.96] & 94.4 (93.3--96.6) & 7.56 & 2.3 \\
\midrule
Stan (posterior mean, 95\,\% CrI) & 1.73 [1.63, 1.84] & 0.54 [0.46, 0.63] & 3.23 [2.75, 3.84] & -- & -- & -- \\
\bottomrule
\end{tabular}
\end{table}

\subsection{Prior knowledge in the conformal calibration}\label{sec:res:prior}
A parameter that is known from other sources is moved from $\bm\theta$ to the known terms $f_0$ of
\eqref{eq:linear_known}. Conformal prediction treats everything that is not estimated as part of the calibration
population. A known parameter can therefore be fixed at its nominal value in the calibration epidemics, or drawn from
a range that reflects how well it is actually known. We show what this implies with a synthetic first and a real case.

\paragraph{Synthetic case: S3I3R.} With all seven rates unknown and $1\,\%$ noise, $\gamma_2$ and $\gamma_3$ are
practically not identifiable, and their uncertainty spills into $\phi_1$ and $\phi_2$ (\cref{tab:s3i3r}). We fix the
recovery rates $\gamma_1, \gamma_2, \gamma_3$ at their nominal values in the estimator, as in \citet{Nielsen2025PIR},
and calibrate with each $\gamma_i$ drawn uniformly within $0$, $\pm 5\,\%$ or $\pm 20\,\%$ of its nominal value. The
coverage is measured on test epidemics with the nominal $\gamma_i$ and with $\gamma_i$ within $\pm 20\,\%$
(\cref{tab:s3:prior}).

Prior knowledge narrows the intervals of the parameters it decouples. With the recovery rates known
exactly, the intervals of $\phi_1$ and $\phi_2$ shrink from $\pm 34\,\%$ and $\pm 33\,\%$ to $\pm 1.6\,\%$, a factor of
twenty, and that of $\beta$ from $\pm 1.8\,\%$ to $\pm 0.7\,\%$.

The precision of the prior knowledge matters. Known only to $\pm 5\,\%$ or $\pm 20\,\%$, the recovery rates still narrow the
intervals of $\phi_1$ and $\phi_2$ ($\pm 9\,\%$ and $\pm 19\,\%$). For $\beta$ and $\theta$, which the data determine
well when all rates are estimated ($\pm 1.8\,\%$ and $\pm 2.4\,\%$), they widen them: to $\pm 4.1\,\%$ and $\pm 17\,\%$ at
$\pm 5\,\%$, and to $\pm 8\,\%$ and $\pm 33\,\%$ at $\pm 20\,\%$. Fixing a parameter that is known less precisely than
the data can determine it is worse than estimating it.

Uncertain inputs belong in the calibration population. Calibrated as exactly known, the intervals
cover $90.9\,\%$ of the test epidemics with the nominal $\gamma_i$, but only $2.4\,\%$ when the $\gamma_i$ deviate by
up to $\pm 20\,\%$: they are narrow intervals around biased estimates. A calibration with $\pm 5\,\%$ gives $59\,\%$
on these test epidemics ($74\,\%$ with PIWR-RACC), and one with $\pm 20\,\%$ gives $89.9\,\%$, the target. When the
$\gamma_i$ are in fact nominal, the wider calibrations are conservative ($100\,\%$). The calibration range must cover
the true uncertainty of the prior knowledge; nothing else in the method changes.

\begin{table}[tb]\centering\scriptsize
\caption{S3I3R with the recovery rates $\gamma_1, \gamma_2, \gamma_3$ fixed at their nominal values in the estimator
(PIWR, $1\,\%$ noise): median half-width (in \% of the actual value) of the 90\,\% simultaneous intervals at the
literature values for different calibrations of $\gamma_i$, and mean coverage (range over five calibration/test splits) on
test epidemics with the nominal $\gamma_i$ and with $\gamma_i$ within $\pm 20\,\%$. First row: all seven rates estimated
(\cref{tab:s3i3r}), coverage on its own calibration population.}
\label{tab:s3:prior}
\setlength{\tabcolsep}{4pt}
\begin{tabular}{l r r r r c c}
\toprule
 & \multicolumn{4}{c}{90\,\% half-width [\%]} & \multicolumn{2}{c}{coverage [\%], test $\gamma_i$}\\
\cmidrule(lr){2-5}\cmidrule(lr){6-7}
$\gamma_1, \gamma_2, \gamma_3$ & $\beta$ & $\phi_1$ & $\phi_2$ & $\theta$ & nominal & within $\pm 20\,\%$\\
\midrule
estimated & 1.8 & 34 & 33 & 2.4 & \multicolumn{2}{c}{89.3 (own population)} \\
known exactly & 0.68 & 1.6 & 1.6 & 2.7 & 90.9 (90.1--92.0) & 2.4 (2.2--2.6) \\
known to $\pm5\,\%$ & 4.1 & 9.4 & 9.6 & 17 & 100.0 (100.0--100.0) & 59.0 (52.6--65.5) \\
known to $\pm20\,\%$ & 8.2 & 19 & 19 & 33 & 100.0 (100.0--100.0) & 89.9 (88.2--90.8) \\
\bottomrule
\end{tabular}
\end{table}

\paragraph{Real case: boarding school.} Suppose the recovery rate is known from clinical experience, $\gamma = 0.5$
per day, an infectious period of about two days. With $\gamma$ fixed, \eqref{eq:school} is linear in $(\beta, I_0)$
alone, and PIWR reduces to a two-column least-squares solve. The estimator uses $\gamma = 0.5$, and the calibration
epidemics use either $\gamma = 0.5$ (known exactly) or $\gamma$ drawn uniformly from $[0.4, 0.6]$ (known approximately);
all other settings are as in \cref{sec:res:extra} (\cref{tab:school:prior}).

The same calibration rule holds. Calibrated as exactly known, the intervals cover $94.8\,\%$ of the test
epidemics with $\gamma = 0.5$, but only $90.5\,\%$ when the true $\gamma$ lies anywhere in $[0.4, 0.6]$; drawing
$\gamma$ from this range in the calibration restores $95.2\,\%$.

The residual-adaptive factor detects a wrong assumption. A wrong $\gamma$ increases the residual of the
fit, and $h$ widens exactly those intervals: with exact calibration, PIWR-RACC keeps $94.2\,\%$ coverage on the
mismatched test epidemics. For the real data $h = 2.4$, the counts fit the model with $\gamma = 0.5$ less well than a typical calibration epidemic, and the PIWR-RACC intervals widen accordingly.

Prior knowledge narrows the intervals only if the estimator can use it. Here fixing $\gamma$ widens the
plain interval for $\beta$ from $[1.15, 2.47]$ to $[1.24, 3.27]$. In the integral form, the noise of the counts
enters the regressors $C = \int I$, $I^2$ and $IC$. With $\gamma$ free, part of this correlated error is absorbed by
$\gamma$. With $\gamma$ fixed, it is carried by $\beta$ alone. In simulations with the true $\gamma$, the standard
deviation of $\hat\beta$ is $0.41$ with $\gamma$ fixed against $0.18$ with $\gamma$ estimated. The conformal
intervals report this rather than crediting the prior with a precision that the estimator does not deliver.

Adding the daily number of convalescent boys, which the data set also records, as a second observed state does not
help either. In simulations with a convalescent stage it leaves the scatter of the estimates unchanged, and the real
convalescent counts are zero for the first four days, which an exponentially distributed stay in bed cannot
produce. A model with a delayed (multi-stage) bed period would be required.

\begin{table}[tb]\centering\scriptsize
\caption{Boarding-school influenza with prior knowledge of the recovery rate: estimates and 95\,\% simultaneous
conformal intervals for $\beta$ and $\mathcal R_0$, and mean coverage (range over five calibration/test splits) on test
epidemics with $\gamma = 0.5$ and with $\gamma \in [0.4, 0.6]$. First row: $\gamma$ estimated (\cref{tab:school}); its
calibration range of $\gamma$ is shown, and its coverage is measured on its own calibration population.}
\label{tab:school:prior}
\setlength{\tabcolsep}{3pt}
\begin{tabular}{l c c c c c c}
\toprule
 & \multicolumn{2}{c}{$\gamma$ used in} & & & \multicolumn{2}{c}{coverage [\%], test $\gamma$}\\
\cmidrule(lr){2-3}\cmidrule(lr){6-7}
Variant & estimator & calibration & $\beta$ [1/day] & $\mathcal R_0$ & $0.5$ & $[0.4, 0.6]$\\
\midrule
$\gamma$ estimated (PIWR-VP-RACC) & -- & $[0.228, 0.684]$ & 1.81 $[1.15, 2.47]$ & 3.96 $[2.97, 4.96]$ & \multicolumn{2}{c}{94.4 (own population)} \\
$\gamma = 0.5$ known, plain & $0.5$ & $0.5$ & 2.25 $[1.24, 3.27]$ & 4.51 $[2.48, 6.54]$ & 94.8 (93.8--95.8) & 90.5 (88.2--92.6) \\
$\gamma = 0.5$ known, RACC & $0.5$ & $0.5$ & 2.25 $[0.925, 3.59]$ & 4.51 $[1.85, 7.17]$ & 95.1 (94.0--96.4) & 94.2 (92.8--95.0) \\
$\gamma = 0.5$ known, plain & $0.5$ & $[0.4, 0.6]$ & 2.25 $[1.14, 3.37]$ & 4.51 $[2.28, 6.74]$ & 97.1 (96.9--97.8) & 95.2 (94.5--95.8) \\
$\gamma = 0.5$ known, RACC & $0.5$ & $[0.4, 0.6]$ & 2.25 $[0.909, 3.6]$ & 4.51 $[1.82, 7.2]$ & 95.6 (94.7--96.2) & 95.1 (94.5--95.7) \\
\bottomrule
\end{tabular}
\end{table}

\subsection{Time-varying transmission: comparison with PINNs}\label{sec:res:tv}
The DINN benchmarks have constant parameters. In a real outbreak the transmission rate changes with interventions,
behaviour and new variants, and estimating $\beta(t)$, or the effective reproduction number
$\mathcal R_t = \beta(t)S(t)/(\delta N)$, from surveillance counts is the central inverse problem of epidemic
monitoring. \citet{Millevoi2024PINN} provide a clean benchmark for this problem with a physics-informed neural network
(PINN) baseline. Their model is the SIR model \eqref{eq:app:sir} with $N = 56\cdot 10^6$, a known removal rate
$\delta = 1/5\,\mathrm{d}^{-1}$ and $I(0) = 1$, and only the number of infected is observed, once per day. Case~1 has
a constant $\beta = 0.6\,\mathrm{d}^{-1}$, Case~2 a synthetic $\beta(t)$ that produces two waves, and Case~3 the
$\beta(t)$ implied by the $\mathcal R_t$ estimates of the first COVID-19 wave in Italy, all over 90 days with Poisson
counts. Case~4 extends Case~2 to 120 days with Gaussian errors of $40\,\%$ coefficient of variation, rounded and
clipped at zero. The errors are relative 2-norms at 90 equispaced times, for $\beta$ or $\mathcal R_t$ also over the
last 70 (Case~4: 100) days, since the first weeks, with a handful of infected, do not identify $\beta$. The case is
relevant for three reasons: the parameter is a function of time, only one of the three states is observed, and the
PINN results come with training times. The reference $\beta(t)$ of Cases~2--4 is given only as a figure. This figure was digitized from Figs.~5, 7 and 8 of \citet{Millevoi2024PINN} and adjusted it so that the simulated $I(t)$ reproduces the published curves (Case~2: peaks of $1.19\cdot 10^5$ on day 40 and $1.18\cdot 10^5$ on day 85; Case~3: a peak of $1.0\cdot 10^6$ on day 47).

\paragraph{From constant to time-varying coefficients.} Three steps put the problem in the form \eqref{eq:linear}.
First, with $\delta$ known and only $I$ observed, $R(t) = \delta C(t)$ with $C(t) = \int_0^t I\,ds$, and
$S = N - I - \delta C$ follows from the data by the cubic-spline quadrature of \cref{sec:piwr}, as for the
boarding school (\cref{sec:res:extra}). Second, the transmission rate is expanded in $K$ cubic B-splines on
equispaced knots, $\beta(t) = \sum_{k=1}^K b_k\varphi_k(t)$, with $K = 24$ for 90 days, $K = 30$ for 120 days and
$K = 1$ for Case~1. The infected compartment then obeys
\begin{equation}\label{eq:tv}
\dot I + \delta I = \sum_{k=1}^K b_k\,\varphi_k(t)\,\frac{S I}{N},
\end{equation}
which is linear in the coefficients $\mathbf b$. PIR evaluates \eqref{eq:tv} at the interior sampling times with the
finite-difference derivative \eqref{eq:fd}. PIWR integrates it from $0$ to $t_j$,
\begin{equation}\label{eq:tv:weak}
\tilde I_j + \delta C_j = I_0 + \sum_{k=1}^K b_k \int_0^{t_j}\varphi_k\,\frac{S I}{N}\,ds,
\end{equation}
with the unknown $I_0$ as in \eqref{eq:piwr:system}. Third, $K$ coefficients are not determined by noisy counts
without some smoothing, so we add the second-difference penalty $\lambda\lVert\mathbf D_2\mathbf b\rVert_2^2$
\citep[a P-spline;][]{Eilers1996PSpline} with $\lambda$ chosen by generalized cross-validation, and weight each row by
$1/\sqrt{\max(\tilde I_j, 1)}$, the standard deviation of a Poisson count. The estimate remains a single linear
least-squares solve. Its covariance $\Sigma_{\mathbf b}$ follows from the sandwich formula of the penalized
estimator, and the pointwise standard error is $\hat\sigma(t) = (\bm\varphi(t)^\top\Sigma_{\mathbf b}\bm\varphi(t))^{1/2}$,
with $\bm\varphi = (\varphi_1, \dots, \varphi_K)$; for $\mathcal R_t$ it is scaled by $\hat S/(\delta N)$.

\paragraph{Conformal bands.} The simultaneous score \eqref{eq:score} becomes a maximum over time,
$s = \max_{t_i \geq 20\,\mathrm{d}} |\hat\beta(t_i) - \beta(t_i)|/(h\,\hat\sigma(t_i))$, over the 90 evaluation times
after day 20, with the residual-adaptive factor $h$ of \eqref{eq:h}. Its calibrated quantile $\hat q$ gives the
$90\,\%$ simultaneous band $\hat\beta(t) \pm \hat q\,h\,\hat\sigma(t)$. The calibration population is the curve
family of each case with every shape parameter (initial level, onset and width of the decline, final level, and the
amplitude, timing and width of the second wave) drawn uniformly within $\pm 30\,\%$; as before, $200$ tuning, $999$
calibration and $2{,}000$ test problems are simulated with the noise model of the case.

\paragraph{Results.} \Cref{tab:tv} compares the errors with the PINN results of \citet[Tables~2--4]{Millevoi2024PINN}
(mean of 10 training runs; ours are medians over 100 noise realizations). In Cases~1--3 PIR and PIWR are more accurate
than both PINN variants for every quantity, mostly by one to two orders of magnitude (the one exception is the
truncation-limited PIR-FD2 estimate of the constant $\beta$), in milliseconds instead of
12--37 minutes of training. The constant $\beta$ of Case~1 is recovered to $6\cdot 10^{-5}$ (PIR-FD6) and
$1\cdot 10^{-4}$ (PIWR), against $3.6\cdot 10^{-4}$ for the better PINN. For the
time-varying cases the full-period error, $0.27$--$0.70$ for the PINNs because they fail in the first weeks, drops
to $0.02$--$0.07$, and after day 20 PIWR reaches $5.0\cdot 10^{-3}$ for $\beta(t)$ and
$2.7\cdot 10^{-3}$ for $\mathcal R_t$, against $1.6\cdot 10^{-2}$ and $6.7\cdot 10^{-2}$ for the better PINN. PIWR
is the most accurate estimator after day 20 in Cases~2 and 3, and it fits the infected curve to $2\cdot 10^{-3}$ and
$5\cdot 10^{-4}$. In Case~4 the $40\,\%$ errors dominate. PIR-FD6 is the most accurate there, with
$\mathcal R_t$ errors of $0.17$ over the whole period and $0.18$ after day 20, against $0.47$--$0.50$ and
$0.31$--$0.33$ for the PINNs. PIWR is comparable
with the PINN ($0.34$ and $0.39$). In the weak form each equation contains one raw sample with its $40\,\%$ error,
while the finite-difference rows, together with the smoothness penalty, average over neighbouring days.

\begin{figure}[tb]
\centering
\includegraphics[width=\textwidth]{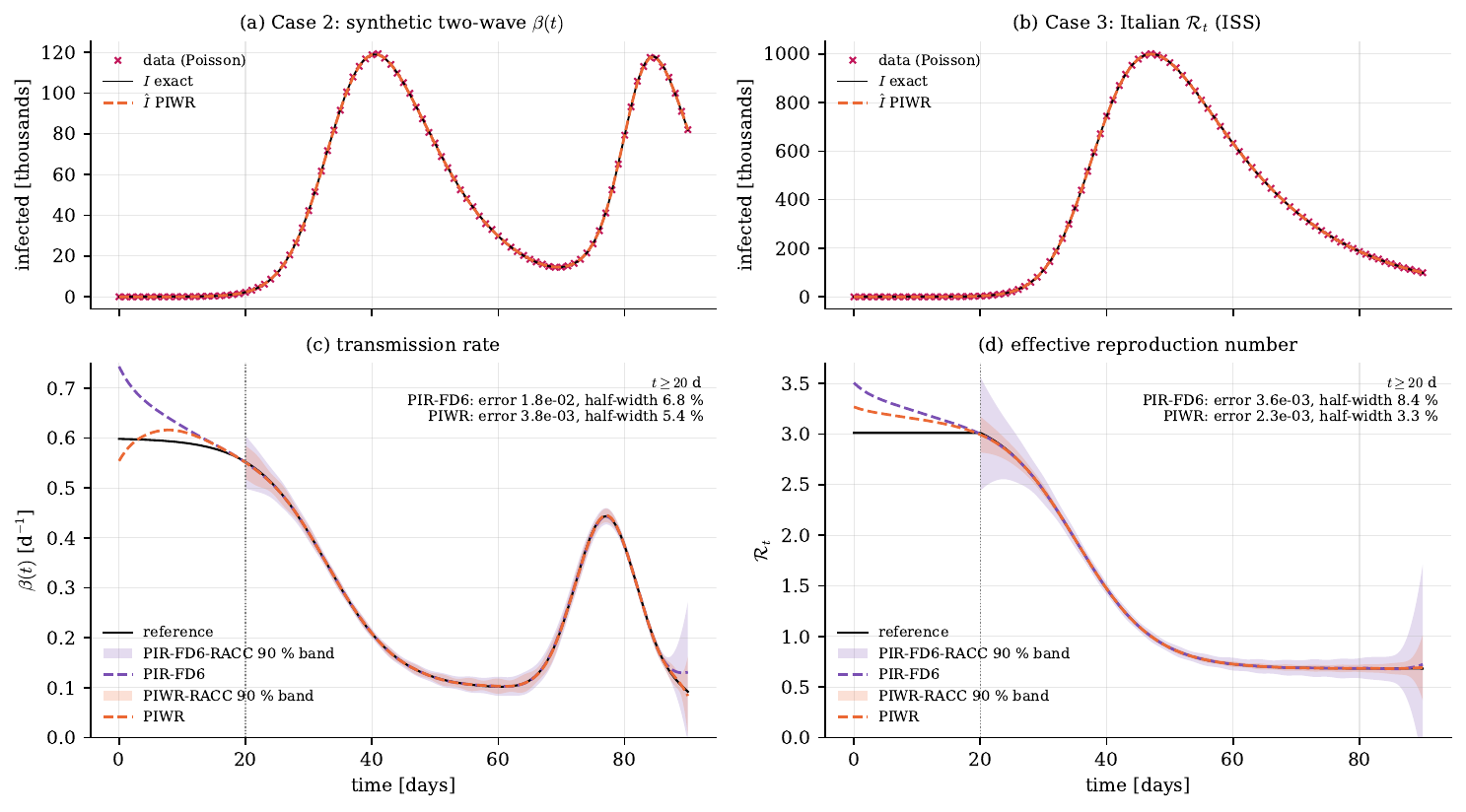}
\caption{Time-varying transmission, Cases~2 and 3 of \citet{Millevoi2024PINN}, one noise realization. (a,~b) Poisson
counts of the infected, exact solution and PIWR fit. (c,~d) Reference $\beta(t)$ and $\mathcal R_t$, the PIR-FD6 and
PIWR estimates, and their $90\,\%$ simultaneous RACC bands after day 20.}
\label{fig:tv}
\end{figure}

\Cref{fig:tv} shows one noise realization of Cases~2 and 3 with the conformal bands. On the test populations the
bands cover $89.2$--$91.9\,\%$ of the curves for both estimators, and they contain the published curves in all of 200
noise realizations, since the population includes sharper transitions than the reference curves. The median
half-width on the reference curves is $5.3\,\%$ (Case~2) and $3.0\,\%$ (Case~3) for PIWR-RACC and $7.1\,\%$ and
$7.2\,\%$ for PIR-FD6-RACC. The residual-adaptive factor is essential in Case~3: problems whose decline the splines
resolve poorly dominate the plain quantile ($\hat q = 191$ for PIWR), and dividing their scores by $h$ lowers it to
$18$, narrowing the bands from $28\,\%$ to $3.0\,\%$. In Case~4 the bands are calibrated ($88.2$--$89.8\,\%$) but
uninformative, with half-widths of $1.7$ (PIR-FD6) and $4.6$ (PIWR) in units of $\mathcal R_t$: with $40\,\%$ errors on
every count, a simultaneous band over 100 days cannot be narrow, and the wide bands correctly reflect that the data do not
determine $\mathcal R_t$ more precisely. Calibration takes
14--27\,s per case and estimator. The PINN bands of \citet{Millevoi2024PINN} are one standard deviation over training
runs, which measures the variability of the optimization on one data set, not the uncertainty due to the data.

\begin{table}[tb]\centering\scriptsize
\caption{Time-varying transmission \citep{Millevoi2024PINN}: relative 2-norm errors at 90 equispaced times and CPU
time per fit for the joint and split PINNs \citep[Tables~2--4]{Millevoi2024PINN} and for PIR-FD2, PIR-FD6 and PIWR
with a B-spline expansion of $\beta(t)$ (medians over 100 noise realizations). $^\dagger$PIR uses the data for $I$;
$S$ and $R$ follow from their integral. The reference curves of Cases 2--4 are digitized (see text); the hardware
differs.}
\label{tab:tv}
\setlength{\tabcolsep}{4pt}
\begin{tabular}{l c c c c c}
\toprule
 & \multicolumn{2}{c}{PINN} & \multicolumn{2}{c}{PIR} & \\
\cmidrule(lr){2-3}\cmidrule(lr){4-5}
Error & joint & split & FD2 & FD6 & PIWR \\
\midrule
\multicolumn{6}{l}{\emph{Case 1: constant $\beta$}} \\
$S$ & $2.8\cdot 10^{-3}$ & $6.2\cdot 10^{-4}$ & $6.1\cdot 10^{-5}$ & $6.1\cdot 10^{-5}$ & $1.6\cdot 10^{-4}$ \\
$I$ & $3.8\cdot 10^{-3}$ & $8.4\cdot 10^{-4}$ & $2.6\cdot 10^{-4}$$^\dagger$ & $2.6\cdot 10^{-4}$$^\dagger$ & $2.0\cdot 10^{-4}$ \\
$R$ & $3.3\cdot 10^{-3}$ & $7.4\cdot 10^{-4}$ & $4.9\cdot 10^{-5}$ & $4.9\cdot 10^{-5}$ & $1.7\cdot 10^{-4}$ \\
$\beta$ & $5.1\cdot 10^{-3}$ & $3.6\cdot 10^{-4}$ & $8.9\cdot 10^{-4}$ & $5.6\cdot 10^{-5}$ & $9.7\cdot 10^{-5}$ \\
CPU time & 2223\,s & 719\,s & 0.15\,ms & 0.14\,ms & 0.22\,ms \\
\midrule
\multicolumn{6}{l}{\emph{Case 2: two-wave $\beta(t)$}} \\
$S$ & $1.8\cdot 10^{-3}$ & $3.4\cdot 10^{-4}$ & $4.8\cdot 10^{-6}$ & $4.8\cdot 10^{-6}$ & $3.6\cdot 10^{-6}$ \\
$I$ & $2.5\cdot 10^{-2}$ & $7.8\cdot 10^{-3}$ & $3.1\cdot 10^{-3}$$^\dagger$ & $3.1\cdot 10^{-3}$$^\dagger$ & $1.8\cdot 10^{-3}$ \\
$R$ & $2.3\cdot 10^{-1}$ & $4.3\cdot 10^{-2}$ & $4.1\cdot 10^{-4}$ & $4.1\cdot 10^{-4}$ & $4.2\cdot 10^{-4}$ \\
$\beta(t)$ & $2.7\cdot 10^{-1}$ & $4.1\cdot 10^{-1}$ & $2.5\cdot 10^{-2}$ & $3.0\cdot 10^{-2}$ & $3.3\cdot 10^{-2}$ \\
$\beta(t)$, last 70 d & $6.0\cdot 10^{-2}$ & $1.6\cdot 10^{-2}$ & $8.2\cdot 10^{-3}$ & $1.2\cdot 10^{-2}$ & $5.0\cdot 10^{-3}$ \\
CPU time & 1936\,s & 717\,s & 4.6\,ms & 4.7\,ms & 6.1\,ms \\
\midrule
\multicolumn{6}{l}{\emph{Case 3: Italian $\mathcal R_t$}} \\
$S$ & $1.9\cdot 10^{-3}$ & $5.0\cdot 10^{-4}$ & $1.4\cdot 10^{-5}$ & $1.4\cdot 10^{-5}$ & $1.1\cdot 10^{-5}$ \\
$I$ & $4.7\cdot 10^{-3}$ & $6.0\cdot 10^{-3}$ & $1.1\cdot 10^{-3}$$^\dagger$ & $1.1\cdot 10^{-3}$$^\dagger$ & $5.4\cdot 10^{-4}$ \\
$R$ & $2.8\cdot 10^{-2}$ & $7.1\cdot 10^{-3}$ & $1.5\cdot 10^{-4}$ & $1.5\cdot 10^{-4}$ & $1.4\cdot 10^{-4}$ \\
$\mathcal R_t$ & $4.7\cdot 10^{-1}$ & $7.0\cdot 10^{-1}$ & $6.5\cdot 10^{-2}$ & $5.3\cdot 10^{-2}$ & $1.7\cdot 10^{-2}$ \\
$\mathcal R_t$, last 70 d & $5.3\cdot 10^{-2}$ & $6.7\cdot 10^{-2}$ & $7.0\cdot 10^{-3}$ & $3.4\cdot 10^{-3}$ & $2.7\cdot 10^{-3}$ \\
CPU time & 1906\,s & 726\,s & 4.8\,ms & 4.6\,ms & 6.2\,ms \\
\midrule
\multicolumn{6}{l}{\emph{Case 4: $40\,\%$ noise, 120 days}} \\
$I$ & $1.4\cdot 10^{-1}$ & $1.3\cdot 10^{-1}$ & $3.6\cdot 10^{-1}$$^\dagger$ & $3.6\cdot 10^{-1}$$^\dagger$ & $2.8\cdot 10^{-1}$ \\
$\mathcal R_t$ & $5.0\cdot 10^{-1}$ & $4.7\cdot 10^{-1}$ & $3.4\cdot 10^{-1}$ & $1.7\cdot 10^{-1}$ & $3.4\cdot 10^{-1}$ \\
$\mathcal R_t$, last 100 d & $3.1\cdot 10^{-1}$ & $3.3\cdot 10^{-1}$ & $2.6\cdot 10^{-1}$ & $1.8\cdot 10^{-1}$ & $3.9\cdot 10^{-1}$ \\
CPU time & 1658\,s & 1053\,s & 6.4\,ms & 6.5\,ms & 7.8\,ms \\
\bottomrule
\end{tabular}
\end{table}

\subsection{Computational cost}\label{sec:res:cost}
A single PIR fit takes $0.14$--$0.5$\,ms and a PIWR fit $0.5$--$1.6$\,ms. The complete conformal pipeline of a model,
including tuning, calibration and testing for all four estimators and both noise levels, takes 19--108\,s
(\cref{tab:uq}). This is dominated by simulating the roughly $3\cdot 10^4$ calibration and test epidemics, and it
parallelizes trivially. The DINN training times reported by \citet{Shaier2022DINN} range from 20 minutes (COVID-19)
to 91 hours (Anthrax) on a 2.3\,GHz Intel Xeon CPU, without uncertainty quantification. The hardware differs, so the
comparison is indicative, but the gap of three to five orders of magnitude is not a matter of hardware.

\subsection{Limitations}\label{sec:res:limits}
\textbf{Sampling.} PIR and PIWR require that the sampled states resolve the dynamics (Anthrax), and that the
  library terms carrying a parameter are not swamped by the noise of the largest compartment. The latter limits, for
  example, the hospital and funeral transmission rates in Ebola.

\textbf{Errors in variables.} Both PIR and PIWR evaluate the library at the noisy samples. Nonlinear library
  terms are then biased (e.g.\ the expected value of $\tilde I^2$ is $I^2 + \sigma^2$), and noise in the regressors
  shrinks the least-squares coefficients. The weak form averages out the random part of this error but not the bias.
  The effect is small at $1\,\%$ noise but dominates for sparse, noisy counts (\cref{sec:res:prior}). The conformal
  calibration includes it, so coverage is retained at the price of wider intervals. We find that smoothing the samples before
  evaluating the library, or bias-corrected or total least squares, could reduce it.

\textbf{Constraints between coefficients.} When $\bm\theta = g(\bm\omega)$ has more entries than $\bm\omega$, a
  free fit ignores the constraint. With few, noisy data this can cost an order of magnitude in accuracy (boarding
  school, \cref{sec:res:extra}). Variable projection restores it when the nonlinearity sits in one or two parameters.
  For sparse count data, a likelihood-based fit is still more efficient than least squares.

\textbf{Scope of the guarantee.} The guarantee is marginal over the calibration population and assumes the model is
  correct. A misspecified model, or a test case far from the prior box, is not covered.

\section{Conclusions}\label{sec:conclusions}

For epidemic compartment models whose right-hand side is linear in coefficients that are parameters or products
of parameters, we conclude that parameter estimation does \emph{not} require training a neural network.
Physics-informed regression solves it with one least-squares problem, and its weak form, PIWR, removes the
dependence on derivative estimates and is fourth-order accurate with cubic-spline quadrature. On the eleven DINN benchmarks \cite{Shaier2022DINN}, PIWR or PIR-FD6 improves on the reported DINN
errors by up to four orders of magnitude wherever the sampling resolves the dynamics, at a cost of milliseconds
instead of hours. Together with the calibrated intervals on real data from the 1978 boarding-school outbreak, where imposing the
coefficient constraint by variable projection (PIWR-VP) was essential with only 15 samples, this
suggests that PIWR is well suited for practical applications whenever the data resolve the epidemic dynamics. However, this does not make other estimation methods obsolete, and therefore comparing with alternative approaches, such as likelihood-based or Bayesian inference, remains valuable as part of a validation procedure. 

The physics-informed regression framework is not restricted to constant parameters. Expanding a time-varying transmission rate in B-splines keeps
the model linear in the coefficients, and a smoothness penalty turns the estimate into one penalized least-squares
solve. On the benchmark of \citet{Millevoi2024PINN}, PIR and PIWR are more accurate than PINNs in three of four cases and at least comparable in the noisiest one, in milliseconds instead of tens of minutes (\cref{sec:res:tv}). The conformal layer extends directly to simultaneous bands for $\beta(t)$ and $\mathcal R_t$, and with $40\,\%$ errors on every count these bands become wide, correctly reflecting that the data do not determine $\mathcal R_t$ more precisely.

A physics-based model can generate its own calibration population, which makes split conformal prediction applicable to model parameters. The resulting simultaneous intervals keep their 90\,\%
coverage in all 88 tested settings of the DINN benchmarks, and in the S3I3R and real-data cases, are narrow where the estimate is accurate, and widen where the data do not
support the estimate, without access to the true parameters. The residual-adaptive factor (PIWR-RACC) adapts the width to
the difficulty of the individual data set at no cost in coverage (in the time-varying Case~3 it narrows the bands by a
factor of nine). Prior knowledge enters the same way, where a parameter known only approximately must be drawn from its uncertainty range in the calibration population, otherwise the intervals lose their coverage, and fixing a parameter narrows the intervals only of the parameters it decouples. 

\section*{Acknowledgment}
This work was performed at Department of Applied Mathematics and Computer Science, Technical University of Denmark.

\section*{Declaration}

The author declare that he had no potential conflicts of interest with respect to the
investigation, authorship, and/or publication of this article.

During the preparation of this work, generative AI tools were used in order to review python code (fx. bug finding), setup scripts for systematic numerical experiments, and improve language clarity. The author take full responsibility for the scientific content of the published article. The original ideas of this work was the outcome of authors human thoughts and creativity.

\section*{Code and hardware environment}

The code that reproduces every figure and table in \cref{sec:results} will be available after peer review of the work. The timings reported in \cref{sec:results} were measured on a MacBook Pro (Nov 2024, Apple M4 Max, 16 cores, 48 GB RAM). The implementation uses Python with NumPy, SciPy and Matplotlib. The ODEs are integrated with LSODA (\texttt{scipy.integrate.odeint}), the weak-form integrals use \texttt{scipy.interpolate.CubicSpline}, and all estimates are dense least-squares solves (\texttt{numpy.linalg.lstsq}) in double precision. No deep-learning framework, automatic differentiation or GPU is required.

\section*{Data}
The benchmark data are regenerated from the generator notebooks of \citet{Shaier2022DINN}
(\url{https://github.com/Shaier/DINN}) with the same initial conditions, time grids and parameters, and agree with
the published data files to about $10^{-8}$ relative difference, except for the Anthrax file (see
\cref{app:anthrax}). The S3I3R data are generated with the parameters of \citet{Nielsen2025PIR} (\cref{app:s3i3r}). The
boarding-school data are \texttt{influenza\_england\_1978\_school} from the R package \texttt{outbreaks}
(\url{https://github.com/reconhub/outbreaks}). The time-varying cases of \cref{sec:res:tv} are generated with the setup of \citet{Millevoi2024PINN};
their reference transmission curves, which are published only as figures, are digitized from Figs.~5, 7 and 8 of
that paper.

\appendix
\section{The eleven DINN epidemic models}\label{app:models}

All data are generated as in the DINN generator notebooks \citep{Shaier2022DINN} (initial conditions, time grids
and parameter values), but integrated with LSODA \citep{Petzold1983LSODA} at relative and absolute tolerances of
$10^{-10}$. Where the generator code differs from the equations printed in \citet{Shaier2022DINN}, the generator is
used, since it produced the benchmark data; the differences are noted below. Every model is stated in the framework
of \cref{sec:models}: the state $\bx$, the equations $\dot\bx = f(\bx, t; \bm\omega)$, the physical parameters
$\bm\omega \in \R^k$ with the values $\bm\omega^\star$ used to generate the data, the coefficient vector
$\bm\theta = g(\bm\omega) \in \R^p$ in which the equations are linear, the left inverse $\bm\omega = h(\bm\theta)$, the
known constants, and the initial condition $\bx(t_0) = \bx_0$ with $t_0 = 0$. The library $\bA(\bx, t)$ of
\eqref{eq:linear} follows by collecting, in each equation, the factor that multiplies each entry of $\bm\theta$.
To avoid a clash with model parameters that are named $\theta$ or $\theta_1$ in the original references, we write
$\vartheta_c$ for the $c$-th entry of $\bm\theta$. When $p > k$, the coefficients satisfy $p - k$ constraints, which
are listed but not imposed. The parameter tables report PIWR with the PIWR-RACC intervals (90\,\%, simultaneous over
all parameters of a model) for noise-free data and for $\epsilon = 1\,\%$, next to the DINN estimates. A dagger
($^\dagger$) marks an absolute error for parameters whose actual value is zero.

\subsection{COVID-19: SIDR model}\label{app:covid}
State $\bx = (S, I, D, R)$ (susceptible, infected, dead, recovered), $n = 4$, with known population
$N = 59\cdot 10^6$:
\begin{equation}
\begin{aligned}
\dot S &= -\alpha\frac{SI}{N},\\
\dot I &= \alpha\frac{SI}{N} - (\beta + \gamma) I,\\
\dot D &= \gamma I,\\
\dot R &= \beta I.
\end{aligned}\label{eq:app:covid}
\end{equation}
Parameters, coefficients and initial condition:
\begin{equation}
\begin{aligned}
\bm\omega &= (\alpha, \beta, \gamma) \in \R^3,\\
\bm\omega^\star &= (0.191,\ 0.05,\ 0.0294),\\
\bm\theta &= g(\bm\omega) = (\alpha, \beta, \gamma) \in \R^3,\\
\bm\omega &= h(\bm\theta) = (\vartheta_1, \vartheta_2, \vartheta_3),\\
\bx_0 &= (N - 1,\ 1,\ 0,\ 0).
\end{aligned}
\end{equation}
The library is given in matrix form in \eqref{eq:sidr}. Data: $m = 100$ samples on $[0, 500]$ days.
\begin{table}[htbp]\centering\scriptsize
\caption{COVID-19 (SIDR): PIWR estimates and PIWR-RACC 90\,\% simultaneous intervals, and the DINN values are from \citet{Shaier2022DINN}.}
\label{tab:app:T01_covid}
\setlength{\tabcolsep}{3pt}
\begin{tabular}{l r r r r r c r r c}
\toprule
 & & \multicolumn{2}{c}{DINN} & \multicolumn{3}{c}{PIWR / PIWR-RACC, $\epsilon = 0$} & \multicolumn{3}{c}{PIWR / PIWR-RACC, $\epsilon = 0.01$}\\
\cmidrule(lr){3-4}\cmidrule(lr){5-7}\cmidrule(lr){8-10}
Par. & Actual & Found & \% err. & Found & \% err. & 90\,\% interval & Found & \% err. & 90\,\% interval\\
\midrule
$\alpha$ & 0.191 & 0.1932 & 1.15 & 0.191 & 0.000105 & [0.191, 0.191] & 0.1908 & 0.124 & [0.1882, 0.1934] \\
$\beta$ & 0.05 & 0.0501 & 0.2 & 0.05 & 0.000108 & [0.05, 0.05] & 0.04976 & 0.487 & [0.04881, 0.05071] \\
$\gamma$ & 0.0294 & 0.0297 & 1.02 & 0.0294 & 7.05e-05 & [0.0294, 0.0294] & 0.02926 & 0.46 & [0.02865, 0.02988] \\
\bottomrule
\end{tabular}
\end{table}

\subsection{HIV}\label{app:hiv}
State $\bx = (T, I, V)$ (healthy CD4$^+$ T cells, infected cells, virus), $n = 3$, with known natural death rate
$\mu_T = 0.02$ and burst factor $N = 250$:
\begin{equation}
\begin{aligned}
\dot T &= s - \mu_T T + rT\Big(1 - \frac{T + I}{T_{\max}} - k_1 V T\Big),\\
\dot I &= k_1' V T - \mu_I I,\\
\dot V &= N\mu_b I - k_1 V T - \mu_V V.
\end{aligned}\label{eq:app:hiv}
\end{equation}
Parameters, coefficients and initial condition:
\begin{equation}
\begin{aligned}
\bm\omega &= (s, \mu_I, \mu_b, \mu_V, r, T_{\max}, k_1, k_1') \in \R^8,\\
\bm\omega^\star &= (10,\ 0.26,\ 0.24,\ 2.4,\ 0.03,\ 1500,\ 2.4\cdot 10^{-4},\ 2\cdot 10^{-4}),\\
\bm\theta &= g(\bm\omega) = (s,\ r - \mu_T,\ r/T_{\max},\ r k_1,\ k_1',\ \mu_I,\ N\mu_b,\ k_1,\ \mu_V) \in \R^9,\\
\bm\omega &= h(\bm\theta) = \big(\vartheta_1,\ \vartheta_6,\ \vartheta_7/N,\ \vartheta_9,\ \vartheta_2 + \mu_T,\ (\vartheta_2 + \mu_T)/\vartheta_3,\ \vartheta_8,\ \vartheta_5\big),\\
\bx_0 &= (1000,\ 10,\ 0.01).
\end{aligned}
\end{equation}
The one constraint is $\vartheta_4 = (\vartheta_2 + \mu_T)\,\vartheta_8$. $\mu_T$ enters only through $r - \mu_T$ and $N$
only through $N\mu_b$, so both are fixed. The generator places $k_1VT$ inside the bracket of the $\dot T$ equation,
unlike the printed equation of \citet{Shaier2022DINN}, and uses $k_1 = 2.4\cdot 10^{-4}$, $k_1' = 2\cdot 10^{-4}$.
Data: $m = 50$ samples on $[0, 20]$ days.
\begin{table}[htbp]\centering\scriptsize
\caption{HIV: PIWR estimates and PIWR-RACC 90\,\% simultaneous intervals, and the DINN values are from \citet{Shaier2022DINN}.}
\label{tab:app:T02_hiv}
\setlength{\tabcolsep}{3pt}
\begin{tabular}{l r r r r r c r r c}
\toprule
 & & \multicolumn{2}{c}{DINN} & \multicolumn{3}{c}{PIWR / PIWR-RACC, $\epsilon = 0$} & \multicolumn{3}{c}{PIWR / PIWR-RACC, $\epsilon = 0.01$}\\
\cmidrule(lr){3-4}\cmidrule(lr){5-7}\cmidrule(lr){8-10}
Par. & Actual & Found & \% err. & Found & \% err. & 90\,\% interval & Found & \% err. & 90\,\% interval\\
\midrule
$s$ & 10 & 10 & 0.00751 & 10.19 & 1.89 & [7.79, 12.59] & 16.77 & 67.7 & [-89.76, 123.3] \\
$\mu_I$ & 0.26 & 0.2613 & 0.489 & 0.2602 & 0.0806 & [0.2592, 0.2612] & 0.2782 & 7 & [0.2313, 0.3251] \\
$\mu_b$ & 0.24 & 0.2417 & 0.728 & 0.2405 & 0.214 & [0.2339, 0.2472] & 0.2545 & 6.04 & [0.04948, 0.4595] \\
$\mu_V$ & 2.4 & 2.42 & 0.83 & 2.404 & 0.181 & [2.345, 2.464] & 2.536 & 5.66 & [0.7169, 4.355] \\
$r$ & 0.03 & 0.03061 & 2.02 & 0.0232 & 22.7 & [-0.05083, 0.09724] & -0.124 & 513 & [-3.335, 3.088] \\
$T_{\max}$ & 1500 & 1507 & 0.436 & 765.8 & 48.9 & [-522, 2054] & 940.1 & 37.3 & [-9800, 1.168e+04] \\
$k_1$ & 0.00024 & 0.000246 & 2.45 & 0.0002579 & 7.44 & [0.0001039, 0.0004118] & 0.0003937 & 64 & [-0.004504, 0.005292] \\
$k_1'$ & 0.0002 & 0.000203 & 1.6 & 0.0002002 & 0.101 & [0.0001993, 0.0002011] & 0.0002134 & 6.68 & [0.0001733, 0.0002534] \\
$\mu_T$ & 0.02 & 0.020762 & 3.81 & \multicolumn{6}{l}{known: only $r-\mu_T$ is identifiable} \\
$N$ & 250 & 249.703 & 0.119 & \multicolumn{6}{l}{known: only $N\mu_b$ is identifiable} \\
\bottomrule
\end{tabular}
\end{table}

\subsection{Smallpox}\label{app:smallpox}
State $\bx = (S, E_n, E_i, C_i, I, Q, U, V)$, $n = 8$, with known population $N = 7000$ and the auxiliary
quantities
\begin{equation}
\begin{aligned}
\varphi &= I/S,\\
\beta &= 10\gamma/(\varphi N).
\end{aligned}\label{eq:app:smallpox:aux}
\end{equation}
The equations are
\begin{equation}
\begin{aligned}
\dot S &= \chi_1(1 - \epsilon_1) C_i - \beta(\varphi + \rho - \varphi\rho) S I,\\
\dot E_n &= \beta\varphi(1 - \rho) S I - \alpha E_n,\\
\dot E_i &= \beta\varphi\rho S I - \big(\chi_1\epsilon_2 + \alpha(1 - \epsilon_2)\big) E_i,\\
\dot C_i &= \beta\rho(1 - \varphi) S I - \chi_1 C_i,\\
\dot I &= \alpha(1 - \theta) E_n - (\theta + \gamma) I,\\
\dot Q &= \alpha(1 - \epsilon_2) E_i + \theta(\alpha E_n + I) - \chi_2 Q,\\
\dot U &= \gamma I + \chi_2 Q,\\
\dot V &= \chi_1(\epsilon_2 E_i + \epsilon_1 C_i).
\end{aligned}\label{eq:app:smallpox}
\end{equation}
With \eqref{eq:app:smallpox:aux}, $\beta\varphi SI = 10\gamma SI/N$ and $\beta SI = 10\gamma S^2/N$, so the equations are
linear in
\begin{equation}
\begin{aligned}
\bm\omega &= (\chi_1, \chi_2, \epsilon_1, \epsilon_2, \rho, \theta, \alpha, \gamma) \in \R^8,\\
\bm\omega^\star &= (0.06,\ 0.04,\ 0.975,\ 0.3,\ 0.975,\ 0.95,\ 0.0685,\ 0.116),\\
\bm\theta &= g(\bm\omega) = (\chi_1,\ \chi_1\epsilon_1,\ \gamma,\ \gamma\rho,\ \alpha,\ \chi_1\epsilon_2,\ \alpha\epsilon_2,\ \theta,\ \alpha\theta,\ \chi_2) \in \R^{10},\\
\bm\omega &= h(\bm\theta) = (\vartheta_1,\ \vartheta_{10},\ \vartheta_2/\vartheta_1,\ \vartheta_6/\vartheta_1,\ \vartheta_4/\vartheta_3,\ \vartheta_8,\ \vartheta_5,\ \vartheta_3),\\
\bx_0 &= (7000,\ 0,\ 0,\ 0,\ 100,\ 0,\ 0,\ 0).
\end{aligned}
\end{equation}
The two constraints are $\vartheta_7 = \vartheta_5\vartheta_6/\vartheta_1$ and $\vartheta_9 = \vartheta_5\vartheta_8$. All eight
parameters are identifiable. The generator uses $\alpha = 0.0685$ and $\gamma = 0.116$ (paper: 0.068 and 0.11).
Data: $m = 50$ samples on $[0, 100]$ days.
\begin{table}[htbp]\centering\scriptsize
\caption{Smallpox: PIWR estimates and PIWR-RACC 90\,\% simultaneous intervals, and the DINN values are from \citet{Shaier2022DINN}.}
\label{tab:app:T03_smallpox}
\setlength{\tabcolsep}{3pt}
\begin{tabular}{l r r r r r c r r c}
\toprule
 & & \multicolumn{2}{c}{DINN} & \multicolumn{3}{c}{PIWR / PIWR-RACC, $\epsilon = 0$} & \multicolumn{3}{c}{PIWR / PIWR-RACC, $\epsilon = 0.01$}\\
\cmidrule(lr){3-4}\cmidrule(lr){5-7}\cmidrule(lr){8-10}
Par. & Actual & Found & \% err. & Found & \% err. & 90\,\% interval & Found & \% err. & 90\,\% interval\\
\midrule
$\chi_1$ & 0.06 & 0.0554 & 7.72 & 0.05598 & 6.7 & [0.03761, 0.07436] & 0.05614 & 6.43 & [0.02666, 0.08563] \\
$\chi_2$ & 0.04 & 0.038 & 4.92 & 0.04017 & 0.418 & [0.02454, 0.05579] & 0.04233 & 5.83 & [0.01734, 0.06732] \\
$\epsilon_1$ & 0.975 & 0.9839 & 0.909 & 1.046 & 7.3 & [0.6453, 1.447] & 1.039 & 6.58 & [0.3989, 1.679] \\
$\epsilon_2$ & 0.3 & 0.2841 & 5.28 & 0.3224 & 7.48 & [-1.355, 2] & 0.1372 & 54.3 & [-2.54, 2.815] \\
$\rho$ & 0.975 & 0.9759 & 0.0882 & 0.9767 & 0.17 & [0.9543, 0.999] & 0.9762 & 0.126 & [0.9396, 1.013] \\
$\theta$ & 0.95 & 0.905 & 4.74 & 0.8834 & 7.01 & [0.2496, 1.517] & 0.797 & 16.1 & [-0.2077, 1.802] \\
$\alpha$ & 0.0685 & 0.0626 & 8.55 & 0.06724 & 1.83 & [0.03579, 0.0987] & 0.06738 & 1.63 & [0.01702, 0.1178] \\
$\gamma$ & 0.116 & 0.1034 & 10.9 & 0.08383 & 27.7 & [0.04272, 0.1249] & 0.08313 & 28.3 & [0.01735, 0.1489] \\
\bottomrule
\end{tabular}
\end{table}

\subsection{Tuberculosis: SLIT model}\label{app:tb}
State $\bx = (S, L, I, T)$ (susceptible, latent, infectious, treated), $n = 4$, with known population $N = 1001$ and
contact rate $c = 1$:
\begin{equation}
\begin{aligned}
\dot S &= \delta - \beta c\frac{SI}{N} - \mu S,\\
\dot L &= \beta c\frac{SI}{N} - (\mu + k + r_1) L + \beta' c\frac{T}{N},\\
\dot I &= k L - (\mu + d) I - r_2 I,\\
\dot T &= r_1 L + r_2 I - \beta' c\frac{T}{N} - \mu T.
\end{aligned}\label{eq:app:tb}
\end{equation}
Parameters, coefficients and initial condition:
\begin{equation}
\begin{aligned}
\bm\omega &= (\delta, \beta, \mu, k, r_1, r_2, \beta', d) \in \R^8,\\
\bm\omega^\star &= (500,\ 13,\ 0.143,\ 0.5,\ 2,\ 1,\ 13,\ 0),\\
\bm\theta &= g(\bm\omega) = (\delta, \beta, \mu, k, r_1, r_2, \beta', d) \in \R^8,\\
\bm\omega &= h(\bm\theta) = (\vartheta_1, \dots, \vartheta_8),\\
\bx_0 &= (1000,\ 0,\ 1,\ 0).
\end{aligned}
\end{equation}
The contact rate enters only through $\beta c$ and $\beta' c$ and is therefore fixed. The actual value $d = 0$ is
reported with an absolute error, and its prior is $U[0, 0.2]$. Data: $m = 50$ samples on $[0, 40]$ days.
\begin{table}[htbp]\centering\scriptsize
\caption{Tuberculosis (SLIT): PIWR estimates and PIWR-RACC 90\,\% simultaneous intervals, and the DINN values are from \citet{Shaier2022DINN}.}
\label{tab:app:T04_tuberculosis}
\setlength{\tabcolsep}{3pt}
\begin{tabular}{l r r r r r c r r c}
\toprule
 & & \multicolumn{2}{c}{DINN} & \multicolumn{3}{c}{PIWR / PIWR-RACC, $\epsilon = 0$} & \multicolumn{3}{c}{PIWR / PIWR-RACC, $\epsilon = 0.01$}\\
\cmidrule(lr){3-4}\cmidrule(lr){5-7}\cmidrule(lr){8-10}
Par. & Actual & Found & \% err. & Found & \% err. & 90\,\% interval & Found & \% err. & 90\,\% interval\\
\midrule
$\delta$ & 500 & 509.5 & 1.86 & 500.2 & 0.0343 & [499.4, 501] & 484.8 & 3.04 & [435.1, 534.5] \\
$\beta$ & 13 & 12.54 & 3.63 & 13.01 & 0.0432 & [12.98, 13.03] & 12.49 & 3.92 & [11, 13.98] \\
$\mu$ & 0.143 & 0.1474 & 3.01 & 0.1429 & 0.037 & [0.1428, 0.1431] & 0.1458 & 1.98 & [0.1335, 0.1582] \\
$k$ & 0.5 & 0.5396 & 7.34 & 0.5001 & 0.0164 & [0.4934, 0.5068] & 0.5489 & 9.77 & [0.125, 0.9728] \\
$r_1$ & 2 & 1.989 & 0.542 & 2.002 & 0.0819 & [1.994, 2.009] & 1.829 & 8.53 & [1.377, 2.281] \\
$r_2$ & 1 & 1.124 & 11.1 & 0.9959 & 0.409 & [0.977, 1.015] & 1.432 & 43.2 & [0.2391, 2.626] \\
$\beta'$ & 13 & 13.74 & 5.37 & 13.04 & 0.325 & [12.92, 13.16] & 10.91 & 16 & [3.29, 18.54] \\
$d$ & 0 & -0.0421 & 0.0421 & 0.004323 & 0.0043$^\dagger$ & [-0.009896, 0.01854] & -0.3259 & 0.33$^\dagger$ & [-1.22, 0.568] \\
$c$ & 1 & 1.0405 & 3.90 & \multicolumn{6}{l}{known: only $\beta c$ and $\beta' c$ are identifiable} \\
\bottomrule
\end{tabular}
\end{table}

\subsection{Pneumonia: SVCIR model}\label{app:pneumonia}
State $\bx = (S, V, C, I, R)$ (susceptible, vaccinated, carrier, infected, recovered), $n = 5$:
\begin{equation}
\begin{aligned}
\dot S &= (1 - p)\pi + \phi V + \delta R - (\mu + \lambda + \theta) S,\\
\dot V &= p\pi + \theta S - (\mu + \epsilon\lambda + \phi) V,\\
\dot C &= \rho\lambda S + \rho\epsilon\lambda V + (1 - q)\eta I - (\mu + \beta + \chi) C,\\
\dot I &= (1 - \rho)\lambda S + (1 - \rho)\epsilon\lambda V + \chi C - (\mu + \alpha + \eta) I,\\
\dot R &= \beta C + q\eta I - (\mu + \delta) R.
\end{aligned}\label{eq:app:pneumonia}
\end{equation}
Parameters, coefficients and initial condition:
\begin{equation}
\begin{aligned}
\bm\omega &= (\pi, \lambda, \epsilon, \phi, \chi, p, \theta, \mu, \alpha, \rho, \beta, \eta, q, \delta) \in \R^{14},\\
\bm\omega^\star &= (0.01,\ 0.1,\ 0.002,\ 0.0025,\ 0.001,\ 0.2,\ 0.008,\ 0.01,\ 0.057,\ 0.05,\ 0.0115,\ 0.2,\ 0.5,\ 0.1),\\
\bm\theta &= g(\bm\omega) = (\pi,\ p\pi,\ \phi,\ \delta,\ \mu,\ \lambda,\ \theta,\ \epsilon\lambda,\ \rho\lambda,\ \rho\epsilon\lambda,\ \chi,\ \beta,\ \alpha,\ \eta,\ q\eta) \in \R^{15},\\
\bm\omega &= h(\bm\theta) = (\vartheta_1,\ \vartheta_6,\ \vartheta_8/\vartheta_6,\ \vartheta_3,\ \vartheta_{11},\ \vartheta_2/\vartheta_1,\ \vartheta_7,\ \vartheta_5,\ \vartheta_{13},\ \vartheta_9/\vartheta_6,\ \vartheta_{12},\ \vartheta_{14},\ \vartheta_{15}/\vartheta_{14},\ \vartheta_4),\\
\bx_0 &= (8200,\ 2800,\ 200,\ 210,\ 200).
\end{aligned}
\end{equation}
The one constraint is $\vartheta_{10} = \vartheta_8\vartheta_9/\vartheta_6$. The parameters $k$ and $\tau$ of the DINN table do
not occur in the model. Data: $m = 50$ samples on $[0, 100]$ days.
\begin{table}[htbp]\centering\scriptsize
\caption{Pneumonia (SVCIR): PIWR estimates and PIWR-RACC 90\,\% simultaneous intervals, and the DINN values are from \citet{Shaier2022DINN}.}
\label{tab:app:T05_pneumonia}
\setlength{\tabcolsep}{3pt}
\begin{tabular}{l r r r r r c r r c}
\toprule
 & & \multicolumn{2}{c}{DINN} & \multicolumn{3}{c}{PIWR / PIWR-RACC, $\epsilon = 0$} & \multicolumn{3}{c}{PIWR / PIWR-RACC, $\epsilon = 0.01$}\\
\cmidrule(lr){3-4}\cmidrule(lr){5-7}\cmidrule(lr){8-10}
Par. & Actual & Found & \% err. & Found & \% err. & 90\,\% interval & Found & \% err. & 90\,\% interval\\
\midrule
$\pi$ & 0.01 & 0.0098 & 2 & 0.06304 & 530 & [-0.116, 0.242] & 1.839 & 1.83e+04 & [-23.99, 27.67] \\
$\lambda$ & 0.1 & 0.099 & 0.962 & 0.09996 & 0.036 & [0.09987, 0.1001] & 0.09767 & 2.33 & [0.08431, 0.111] \\
$\epsilon$ & 0.002 & 0.0022 & 11.7 & 0.002238 & 11.9 & [0.001662, 0.002813] & 0.0204 & 920 & [-0.0651, 0.1059] \\
$\phi$ & 0.0025 & 0.0027 & 7.49 & 0.002465 & 1.4 & [0.002362, 0.002568] & -0.002245 & 190 & [-0.01684, 0.01235] \\
$\chi$ & 0.001 & 0.0011 & 6.74 & 0.000959 & 4.1 & [0.0007445, 0.001173] & -0.002748 & 375 & [-0.03821, 0.03272] \\
$p$ & 0.2 & 0.2033 & 1.64 & 0.1211 & 39.5 & [-1.267, 1.509] & -2.056 & 1.13e+03 & [-33.32, 29.21] \\
$\theta$ & 0.008 & 0.0084 & 4.89 & 0.008003 & 0.0334 & [0.007967, 0.008038] & 0.006661 & 16.7 & [0.001711, 0.01161] \\
$\mu$ & 0.01 & 0.0092 & 8.45 & 0.01002 & 0.153 & [0.009973, 0.01006] & 0.01046 & 4.56 & [0.004406, 0.01651] \\
$\alpha$ & 0.057 & 0.057 & 0.0005 & 0.0569 & 0.172 & [0.05666, 0.05714] & 0.05516 & 3.22 & [0.0209, 0.08943] \\
$\rho$ & 0.05 & 0.0508 & 1.52 & 0.05049 & 0.978 & [0.04934, 0.05164] & 0.06605 & 32.1 & [-0.1241, 0.2563] \\
$\beta$ & 0.0115 & 0.0122 & 5.82 & 0.0115 & 0.0373 & [0.01149, 0.01152] & 0.01177 & 2.38 & [0.009418, 0.01413] \\
$\eta$ & 0.2 & 0.2023 & 1.14 & 0.1999 & 0.031 & [0.1997, 0.2001] & 0.1986 & 0.683 & [0.1698, 0.2275] \\
$q$ & 0.5 & 0.496 & 0.8 & 0.5003 & 0.0541 & [0.4998, 0.5008] & 0.4954 & 0.925 & [0.4228, 0.568] \\
$\delta$ & 0.1 & 0.1038 & 3.75 & 0.1 & 0.0121 & [0.09993, 0.1001] & 0.09902 & 0.984 & [0.08697, 0.1111] \\
$k$ & 0.5 & 0.5025 & 0.508 & \multicolumn{6}{l}{not in the model} \\
$\tau$ & 0.89 & 0.8912 & 0.131 & \multicolumn{6}{l}{not in the model} \\
\bottomrule
\end{tabular}
\end{table}

\subsection{Ebola: SEIHFR model}\label{app:ebola}
State $\bx = (S, E, I, H, F, R)$ (susceptible, exposed, infectious, hospitalized, dead but not yet buried, removed),
$n = 6$, with known population $N = 470\,000$ and known probabilities $\theta_1 = 0.65$, $\delta_1 = 0.47$,
$\delta_2 = 0.42$:
\begin{equation}
\begin{aligned}
\dot S &= -\frac{1}{N}\big(\beta_1 S I + \beta_h S H + \beta_f S F\big),\\
\dot E &= \frac{1}{N}\big(\beta_1 S I + \beta_h S H + \beta_f S F\big) - \alpha E,\\
\dot I &= \alpha E - \big(\gamma_h\theta_1 + \gamma_i(1 - \theta_1)(1 - \delta_1) + \gamma_d(1 - \theta_1)\delta_1\big) I,\\
\dot H &= \gamma_h\theta_1 I - \big(\gamma_{dh}\delta_2 + \gamma_{ih}(1 - \delta_2)\big) H,\\
\dot F &= \gamma_d(1 - \theta_1)\delta_1 I + \gamma_{dh}\delta_2 H - \gamma_f F,\\
\dot R &= \gamma_i(1 - \theta_1)(1 - \delta_1) I + \gamma_{ih}(1 - \delta_2) H + \gamma_f F.
\end{aligned}\label{eq:app:ebola}
\end{equation}
Parameters, coefficients and initial condition:
\begin{equation}
\begin{aligned}
\bm\omega &= (\beta_1, \beta_h, \beta_f, \alpha, \gamma_h, \gamma_i, \gamma_d, \gamma_f, \gamma_{ih}, \gamma_{dh}) \in \R^{10},\\
\bm\omega^\star &= (3.532,\ 0.012,\ 0.462,\ 1/12,\ 1/4.2,\ 1/10,\ 1/8,\ 1/2,\ 1/12.2,\ 1/14.2),\\
\bm\theta &= g(\bm\omega) = (\beta_1, \beta_h, \beta_f, \alpha, \gamma_h, \gamma_i, \gamma_d, \gamma_f, \gamma_{ih}, \gamma_{dh}) \in \R^{10},\\
\bm\omega &= h(\bm\theta) = (\vartheta_1, \dots, \vartheta_{10}),\\
\bx_0 &= (N - 425,\ 0,\ 425,\ 0,\ 0,\ 0).
\end{aligned}
\end{equation}
The probabilities $\theta_1, \delta_1, \delta_2$ enter only as factors of the rates, so they are fixed. The values
$\gamma_{ih}^\star = 1/(1/\gamma_d^\star + 1/\gamma_h^\star)$ and $\gamma_{dh}^\star = 1/(1/\gamma_i^\star + 1/\gamma_h^\star)$
follow the generator. Data: $m = 100$ samples on $[0, 100]$ days.
\begin{table}[htbp]\centering\scriptsize
\caption{Ebola (SEIHFR): PIWR estimates and PIWR-RACC 90\,\% simultaneous intervals, and the DINN values are from \citet{Shaier2022DINN}.}
\label{tab:app:T06_ebola}
\setlength{\tabcolsep}{3pt}
\begin{tabular}{l r r r r r c r r c}
\toprule
 & & \multicolumn{2}{c}{DINN} & \multicolumn{3}{c}{PIWR / PIWR-RACC, $\epsilon = 0$} & \multicolumn{3}{c}{PIWR / PIWR-RACC, $\epsilon = 0.01$}\\
\cmidrule(lr){3-4}\cmidrule(lr){5-7}\cmidrule(lr){8-10}
Par. & Actual & Found & \% err. & Found & \% err. & 90\,\% interval & Found & \% err. & 90\,\% interval\\
\midrule
$\beta_1$ & 3.532 & 3.559 & 0.762 & 3.53 & 0.0598 & [3.526, 3.534] & 4.182 & 18.4 & [2.779, 5.585] \\
$\beta_h$ & 0.012 & 0.0129 & 7.81 & -0.02367 & 297 & [-0.08456, 0.03722] & 0.5563 & 4.54e+03 & [-12.45, 13.57] \\
$\beta_f$ & 0.462 & 0.4638 & 0.398 & 0.8687 & 88 & [0.17, 1.567] & -23.29 & 5.14e+03 & [-187.1, 140.5] \\
$\alpha$ & 0.08333 & 0.0866 & 3.93 & 0.08333 & 3.41e-05 & [0.08333, 0.08333] & 0.08334 & 0.0134 & [0.08032, 0.08637] \\
$\gamma_h$ & 0.2381 & 0.2471 & 3.79 & 0.2381 & 9.95e-06 & [0.2381, 0.2381] & 0.2379 & 0.0721 & [0.2243, 0.2516] \\
$\gamma_i$ & 0.1 & 0.0904 & 9.64 & 0.1 & 0.000556 & [0.09999, 0.1] & 0.1226 & 22.6 & [0.05998, 0.1853] \\
$\gamma_d$ & 0.125 & 0.1205 & 3.61 & 0.125 & 0.000133 & [0.125, 0.125] & 0.1061 & 15.1 & [0.07988, 0.1324] \\
$\gamma_f$ & 0.5 & 0.5196 & 3.92 & 0.5 & 0.000314 & [0.4999, 0.5001] & 0.3156 & 36.9 & [0.05221, 0.5789] \\
$\gamma_{ih}$ & 0.08197 & 0.0811 & 1.09 & 0.08197 & 0.000215 & [0.08196, 0.08197] & 0.1048 & 27.8 & [0.07144, 0.1381] \\
$\gamma_{dh}$ & 0.07042 & 0.071 & 0.756 & 0.07042 & 0.000378 & [0.07041, 0.07043] & 0.03879 & 44.9 & [-0.006073, 0.08366] \\
$\theta_1$ & 0.65 & 0.6523 & 0.348 & \multicolumn{6}{l}{known: only rate products are identifiable} \\
$\delta_1$ & 0.47 & 0.4712 & 0.257 & \multicolumn{6}{l}{known: only rate products are identifiable} \\
$\delta_2$ & 0.42 & 0.4247 & 1.12 & \multicolumn{6}{l}{known: only rate products are identifiable} \\
\bottomrule
\end{tabular}
\end{table}

\subsection{Dengue: host--vector model}\label{app:dengue}
State $\bx = (S_h, E_h, I_h, R_h, S_v, E_v, I_v)$ (hosts $h$, vectors $v$), $n = 7$:
\begin{equation}
\begin{aligned}
\dot S_h &= \pi_h - \lambda_h S_h - \mu_h S_h,\\
\dot E_h &= \lambda_h S_h - \sigma_h\mu_h E_h,\\
\dot I_h &= \sigma_h E_h - (\tau_h + \mu_h + \delta_h) I_h,\\
\dot R_h &= \tau_h I_h - \mu_h R_h,\\
\dot S_v &= \pi_v - \delta_v S_v - \mu_v S_v,\\
\dot E_v &= \delta_v S_v - (\sigma_v + \mu_v) E_v,\\
\dot I_v &= \sigma_v E_v - (\mu_v + \delta_v) I_v.
\end{aligned}\label{eq:app:dengue}
\end{equation}
Parameters, coefficients and initial condition:
\begin{equation}
\begin{aligned}
\bm\omega &= (\pi_h, \pi_v, \lambda_h, \delta_h, \delta_v, \mu_h, \mu_v, \sigma_h, \sigma_v, \tau_h) \in \R^{10},\\
\bm\omega^\star &= (10,\ 30,\ 0.0553,\ 0.99,\ 0.057,\ 0.0195,\ 0.016,\ 0.53,\ 0.2,\ 0.1),\\
\bm\theta &= g(\bm\omega) = (\pi_h,\ \lambda_h,\ \mu_h,\ \sigma_h\mu_h,\ \sigma_h,\ \tau_h,\ \delta_h,\ \pi_v,\ \delta_v,\ \mu_v,\ \sigma_v) \in \R^{11},\\
\bm\omega &= h(\bm\theta) = (\vartheta_1,\ \vartheta_8,\ \vartheta_2,\ \vartheta_7,\ \vartheta_9,\ \vartheta_3,\ \vartheta_{10},\ \vartheta_5,\ \vartheta_{11},\ \vartheta_6),\\
\bx_0 &= (100,\ 0,\ 0,\ 0,\ 1,\ 0,\ 0).
\end{aligned}
\end{equation}
The one constraint is $\vartheta_4 = \vartheta_3\vartheta_5$. The parameter $\lambda_v$ of the DINN table does not occur in
the model, and the generator uses $\lambda_h = 0.0553$. Data: $m = 50$ samples on $[0, 200]$ days.
\begin{table}[htbp]\centering\scriptsize
\caption{Dengue: PIWR estimates and PIWR-RACC 90\,\% simultaneous intervals, and the DINN values are from \citet{Shaier2022DINN}.}
\label{tab:app:T07_dengue}
\setlength{\tabcolsep}{3pt}
\begin{tabular}{l r r r r r c r r c}
\toprule
 & & \multicolumn{2}{c}{DINN} & \multicolumn{3}{c}{PIWR / PIWR-RACC, $\epsilon = 0$} & \multicolumn{3}{c}{PIWR / PIWR-RACC, $\epsilon = 0.01$}\\
\cmidrule(lr){3-4}\cmidrule(lr){5-7}\cmidrule(lr){8-10}
Par. & Actual & Found & \% err. & Found & \% err. & 90\,\% interval & Found & \% err. & 90\,\% interval\\
\midrule
$\pi_h$ & 10 & 9.932 & 0.683 & 10 & 0.000997 & [9.992, 10.01] & 9.964 & 0.357 & [6.964, 12.96] \\
$\pi_v$ & 30 & 29.85 & 0.486 & 30 & 0.00301 & [29.97, 30.03] & 29.85 & 0.499 & [20.53, 39.17] \\
$\lambda_h$ & 0.0553 & 0.0552 & 0.27 & 0.0553 & 1.24e-05 & [0.05524, 0.05536] & 0.0558 & 0.907 & [0.03434, 0.07727] \\
$\delta_h$ & 0.99 & 0.9643 & 2.6 & 0.9919 & 0.194 & [0.9895, 0.9944] & 0.5389 & 45.6 & [-0.009542, 1.087] \\
$\delta_v$ & 0.057 & 0.0567 & 0.529 & 0.057 & 0.00302 & [0.05695, 0.05705] & 0.05665 & 0.606 & [0.0393, 0.07401] \\
$\mu_h$ & 0.0195 & 0.0194 & 0.384 & 0.0195 & 0.00383 & [0.01948, 0.01952] & 0.01892 & 2.98 & [0.01269, 0.02515] \\
$\mu_v$ & 0.016 & 0.0159 & 0.88 & 0.016 & 0.00342 & [0.01598, 0.01602] & 0.01577 & 1.45 & [0.009699, 0.02184] \\
$\sigma_h$ & 0.53 & 0.5372 & 1.36 & 0.5309 & 0.172 & [0.5297, 0.5321] & 0.3142 & 40.7 & [0.05427, 0.5742] \\
$\sigma_v$ & 0.2 & 0.1989 & 0.548 & 0.2 & 0.00324 & [0.1998, 0.2002] & 0.1992 & 0.424 & [0.1385, 0.2598] \\
$\tau_h$ & 0.1 & 0.0902 & 9.77 & 0.1 & 0.00271 & [0.09994, 0.1001] & 0.09768 & 2.32 & [0.07686, 0.1185] \\
$\lambda_v$ & 0.05 & 0.0506 & 1.29 & \multicolumn{6}{l}{not in the model} \\
\bottomrule
\end{tabular}
\end{table}

\subsection{Anthrax: SIAC model}\label{app:anthrax}
State $\bx = (S, I, A, C)$ (susceptible and infected animals, spores in the environment, carcasses), $n = 4$:
\begin{equation}
\begin{aligned}
\dot S &= r(S + I)\Big(1 - \frac{S + I}{K}\Big) - \eta_a A S - \eta_c S C - \eta_i\frac{S I}{S + I} - \mu S + \tau I,\\
\dot I &= \eta_a A S + \eta_c S C + \Big(\eta_i\frac{S I}{S + I} - (\gamma + \mu + \tau)\Big) I,\\
\dot A &= -\sigma A + \beta C,\\
\dot C &= (\gamma + \mu) I - \delta(S + I) C - \kappa C.
\end{aligned}\label{eq:app:anthrax}
\end{equation}
Parameters, coefficients and initial condition:
\begin{equation}
\begin{aligned}
\bm\omega &= (r, \mu, \kappa, \eta_a, \eta_c, \eta_i, \tau, \gamma, \delta, K, \beta, \sigma) \in \R^{12},\\
\bm\omega^\star &= (1/300,\ 1/600,\ 1/10,\ 1/2,\ 1/10,\ 1/100,\ 1/10,\ 1/7,\ 1/20,\ 100,\ 1/500,\ 1/10),\\
\bm\theta &= g(\bm\omega) = (r,\ r/K,\ \mu,\ \kappa,\ \eta_a,\ \eta_c,\ \eta_i,\ \tau,\ \gamma,\ \delta,\ \beta,\ \sigma) \in \R^{12},\\
\bm\omega &= h(\bm\theta) = (\vartheta_1,\ \vartheta_3,\ \vartheta_4,\ \vartheta_5,\ \vartheta_6,\ \vartheta_7,\ \vartheta_8,\ \vartheta_9,\ \vartheta_{10},\ \vartheta_1/\vartheta_2,\ \vartheta_{11},\ \vartheta_{12}),\\
\bx_0 &= (20,\ 3,\ 0,\ 0).
\end{aligned}
\end{equation}
The generator uses $\delta = 1/20$ and $\beta = 1/500$ (paper table: $1/64$ and $0.02$). The DINN data file was
produced with the default LSODA tolerances and differs from the accurate solution by up to 24\,\%. Data: $m = 100$
samples on $[0, 4000]$ days.
\begin{table}[htbp]\centering\scriptsize
\caption{Anthrax (SIAC): PIWR estimates and PIWR-RACC 90\,\% simultaneous intervals, and the DINN values are from \citet{Shaier2022DINN}.}
\label{tab:app:T08_anthrax}
\setlength{\tabcolsep}{3pt}
\begin{tabular}{l r r r r r c r r c}
\toprule
 & & \multicolumn{2}{c}{DINN} & \multicolumn{3}{c}{PIWR / PIWR-RACC, $\epsilon = 0$} & \multicolumn{3}{c}{PIWR / PIWR-RACC, $\epsilon = 0.01$}\\
\cmidrule(lr){3-4}\cmidrule(lr){5-7}\cmidrule(lr){8-10}
Par. & Actual & Found & \% err. & Found & \% err. & 90\,\% interval & Found & \% err. & 90\,\% interval\\
\midrule
$r$ & 0.003333 & 0.0034 & 1.2 & -0.1327 & 4.08e+03 & [-15.16, 14.89] & -0.1364 & 4.19e+03 & [-0.4602, 0.1873] \\
$\mu$ & 0.001667 & 0.0017 & 0.575 & -0.1343 & 8.16e+03 & [-15.16, 14.89] & -0.1381 & 8.39e+03 & [-0.4619, 0.1856] \\
$\kappa$ & 0.1 & 0.1025 & 2.54 & 0.0409 & 59.1 & [-30.33, 30.42] & 0.03519 & 64.8 & [-0.7564, 0.8268] \\
$\eta_a$ & 0.5 & 0.5035 & 0.702 & -1.957 & 491 & [-582, 578.1] & 0.4914 & 1.72 & [-12.2, 13.18] \\
$\eta_c$ & 0.1 & 0.1024 & 2.45 & 0.06861 & 31.4 & [-9.067, 9.204] & 0.02374 & 76.3 & [-0.1683, 0.2157] \\
$\eta_i$ & 0.01 & 0.0106 & 6.05 & 0.01014 & 1.39 & [-3.771, 3.791] & 0.006607 & 33.9 & [-0.09789, 0.1111] \\
$\tau$ & 0.1 & 0.0976 & 2.45 & 0.07203 & 28 & [-18.83, 18.97] & 0.07272 & 27.3 & [-0.3677, 0.5131] \\
$\gamma$ & 0.1429 & 0.1444 & 1.05 & 0.1424 & 0.314 & [-1.634, 1.919] & 0.143 & 0.124 & [0.08593, 0.2001] \\
$\delta$ & 0.05 & 0.0512 & 2.35 & -0.001077 & 102 & [-5.542, 5.54] & -0.001847 & 104 & [-0.1247, 0.121] \\
$K$ & 100 & 100.6 & 0.639 & -4059 & 4.16e+03 & [-8.01e+05, 7.928e+05] & -4226 & 4.33e+03 & [-2.641e+04, 1.795e+04] \\
$\beta$ & 0.002 & 0.0021 & 6.55 & 0.001881 & 5.95 & [-67.99, 68] & 0.001094 & 45.3 & [-1.566, 1.568] \\
$\sigma$ & 0.1 & 0.1051 & 5.1 & 0.09382 & 6.18 & [-3384, 3384] & 0.05752 & 42.5 & [-82.01, 82.13] \\
\bottomrule
\end{tabular}
\end{table}

\subsection{Polio: age-structured SIR model}\label{app:polio}
State $\bx = (S_c, S_a, I_c, I_a, R_c, R_a)$ (children $c$, adults $a$), $n = 6$, with the group sizes fixed by the
initial condition,
\begin{equation}
\begin{aligned}
N_c &= S_c(0) + I_c(0) + R_c(0) = 91.2,\\
N_a &= S_a(0) + I_a(0) + R_a(0) = 908.8,\\
N &= N_c + N_a = 1000.
\end{aligned}\label{eq:app:polio:N}
\end{equation}
The equations are
\begin{equation}
\begin{aligned}
\dot S_c &= \mu N - \Big(\alpha + \mu + \frac{\beta_{cc}}{N_c} I_c + \frac{\beta_{ca}}{N_c} I_a\Big) S_c,\\
\dot S_a &= \alpha S_c - \Big(\mu + \frac{\beta_{aa}}{N_a} I_a + \frac{\beta_{ac}}{N_a} I_c\Big) S_a,\\
\dot I_c &= \Big(\frac{\beta_{cc}}{N_c} I_c + \frac{\beta_{ca}}{N_c} I_a\Big) S_c - (\gamma_c + \alpha + \mu) I_c,\\
\dot I_a &= \Big(\frac{\beta_{ac}}{N_a} I_c + \frac{\beta_{aa}}{N_a} I_a\Big) S_a - (\gamma_a + \mu) I_a + \alpha I_c,\\
\dot R_c &= \gamma_c I_c - \mu R_c - \alpha R_c,\\
\dot R_a &= \gamma_a I_a - \mu R_a + \alpha R_c.
\end{aligned}\label{eq:app:polio}
\end{equation}
Parameters, coefficients and initial condition:
\begin{equation}
\begin{aligned}
\bm\omega &= (\mu, \alpha, \gamma_a, \gamma_c, \beta_{aa}, \beta_{cc}, \beta_{ac}, \beta_{ca}) \in \R^8,\\
\bm\omega^\star &= (0.02,\ 0.5,\ 18,\ 36,\ 40,\ 90,\ 0,\ 0),\\
\bm\theta &= g(\bm\omega) = (\mu, \alpha, \gamma_a, \gamma_c, \beta_{aa}, \beta_{cc}, \beta_{ac}, \beta_{ca}) \in \R^8,\\
\bm\omega &= h(\bm\theta) = (\vartheta_1, \dots, \vartheta_8),\\
\bx_0 &= (35.5,\ 355,\ 0.3,\ 0,\ 55.4,\ 553.8).
\end{aligned}
\end{equation}
The cross rates $\beta_{ac} = \beta_{ca} = 0$ have the prior $U[0, 2]$ and are reported with absolute errors. Data:
$m = 50$ samples on $[0, 40]$ days.
\begin{table}[htbp]\centering\scriptsize
\caption{Polio: PIWR estimates and PIWR-RACC 90\,\% simultaneous intervals, and the DINN values are from \citet{Shaier2022DINN}.}
\label{tab:app:T09_polio}
\setlength{\tabcolsep}{3pt}
\begin{tabular}{l r r r r r c r r c}
\toprule
 & & \multicolumn{2}{c}{DINN} & \multicolumn{3}{c}{PIWR / PIWR-RACC, $\epsilon = 0$} & \multicolumn{3}{c}{PIWR / PIWR-RACC, $\epsilon = 0.01$}\\
\cmidrule(lr){3-4}\cmidrule(lr){5-7}\cmidrule(lr){8-10}
Par. & Actual & Found & \% err. & Found & \% err. & 90\,\% interval & Found & \% err. & 90\,\% interval\\
\midrule
$\mu$ & 0.02 & 0.02 & 0.02 & 0.02001 & 0.0359 & [0.01869, 0.02132] & 0.02006 & 0.29 & [0.01709, 0.02303] \\
$\alpha$ & 0.5 & 0.5018 & 0.36 & 0.5002 & 0.0361 & [0.4681, 0.5323] & 0.5027 & 0.542 & [0.43, 0.5754] \\
$\gamma_a$ & 18 & 18.02 & 0.417 & 18.01 & 0.0469 & [16.83, 19.19] & 18.07 & 0.364 & [15.42, 20.71] \\
$\gamma_c$ & 36 & 36.07 & 0.359 & 36 & 0.01 & [29.2, 42.81] & 33.28 & 7.56 & [18.2, 48.35] \\
$\beta_{aa}$ & 40 & 40.25 & 0.627 & 40.01 & 0.0207 & [37.08, 42.94] & 40.59 & 1.48 & [34.08, 47.11] \\
$\beta_{cc}$ & 90 & 90.61 & 0.672 & 89.21 & 0.875 & [7.781, 170.6] & 83.41 & 7.32 & [-96.75, 263.6] \\
$\beta_{ac}$ & 0 & 0.0002 & 0.0002 & 0.2758 & 0.28$^\dagger$ & [-81.54, 82.09] & -14.52 & 15$^\dagger$ & [-196, 166.9] \\
$\beta_{ca}$ & 0 & 0.0004 & 0.0004 & 0.02784 & 0.028$^\dagger$ & [-2.862, 2.917] & -0.08872 & 0.089$^\dagger$ & [-6.469, 6.291] \\
\bottomrule
\end{tabular}
\end{table}

\subsection{Measles: seasonally forced SEI model}\label{app:measles}
State $\bx = (S, E, I)$, $n = 3$, with known population $N = 50\,001$ and the seasonally forced transmission rate
\begin{equation}
\beta(t) = \beta_0\big(1 + \beta_1\cos 2\pi t\big). \label{eq:app:measles:beta}
\end{equation}
The equations are
\begin{equation}
\begin{aligned}
\dot S &= \mu(N - S) - \beta(t)\frac{S I}{N},\\
\dot E &= \beta(t)\frac{S I}{N} - \mu\sigma E,\\
\dot I &= \sigma E - (\mu + \gamma) I.
\end{aligned}\label{eq:app:measles}
\end{equation}
Parameters, coefficients and initial condition:
\begin{equation}
\begin{aligned}
\bm\omega &= (\mu, \beta_0, \beta_1, \gamma, \sigma) \in \R^5,\\
\bm\omega^\star &= (0.02,\ 0.0010107\,N,\ 0.28,\ 100,\ 35.84),\\
\bm\theta &= g(\bm\omega) = (\mu,\ \beta_0,\ \beta_0\beta_1,\ \mu\sigma,\ \sigma,\ \gamma) \in \R^6,\\
\bm\omega &= h(\bm\theta) = (\vartheta_1,\ \vartheta_2,\ \vartheta_3/\vartheta_2,\ \vartheta_6,\ \vartheta_5),\\
\bx_0 &= (50\,000,\ 0,\ 1).
\end{aligned}
\end{equation}
The one constraint is $\vartheta_4 = \vartheta_1\vartheta_5$. The known forcing $\cos 2\pi t$ enters the library $A(\bx, t)$,
which is the reason for its explicit time dependence. Time is measured in years (the forcing has period one).
DINN treats $\beta_0^\star = 0.0010107\,N \approx 50.54$ as known; here it is estimated. Data: $m = 50$ samples on
$[0, 10]$.
\begin{table}[htbp]\centering\scriptsize
\caption{Measles (seasonal SEI): PIWR estimates and PIWR-RACC 90\,\% simultaneous intervals, and the DINN values are from \citet{Shaier2022DINN}.}
\label{tab:app:T10_measles}
\setlength{\tabcolsep}{3pt}
\begin{tabular}{l r r r r r c r r c}
\toprule
 & & \multicolumn{2}{c}{DINN} & \multicolumn{3}{c}{PIWR / PIWR-RACC, $\epsilon = 0$} & \multicolumn{3}{c}{PIWR / PIWR-RACC, $\epsilon = 0.01$}\\
\cmidrule(lr){3-4}\cmidrule(lr){5-7}\cmidrule(lr){8-10}
Par. & Actual & Found & \% err. & Found & \% err. & 90\,\% interval & Found & \% err. & 90\,\% interval\\
\midrule
$\mu$ & 0.02 & 0.0225 & 12.7 & 0.01834 & 8.31 & [0.01411, 0.02257] & 0.01882 & 5.92 & [-0.07752, 0.1152] \\
$\beta_0$ & 50.54 & -- & -- & 47.5 & 6.01 & [43.23, 51.77] & 48.66 & 3.71 & [-30.79, 128.1] \\
$\beta_1$ & 0.28 & 0.27 & 3.57 & 0.09722 & 65.3 & [-0.2567, 0.4511] & 0.2006 & 28.4 & [-7.031, 7.432] \\
$\gamma$ & 100 & 97 & 3 & 90.61 & 9.39 & [75.66, 105.6] & 29.45 & 70.5 & [-203.7, 262.6] \\
$\sigma$ & 35.84 & 34.71 & 3.15 & 32.46 & 9.44 & [27.09, 37.82] & 10.53 & 70.6 & [-73.46, 94.52] \\
\bottomrule
\end{tabular}
\end{table}

\subsection{Zika: host--vector model}\label{app:zika}
State $\bx = (S_h, E_h, I_{h1}, I_{h2}, A_h, R_h, S_v, E_v, I_v)$ (hosts $h$ with two infectious stages and
asymptomatic $A_h$, vectors $v$), $n = 9$, with known biting rate $a = 0.5$, host and vector populations fixed by the
initial condition, and the forces of infection on the hosts from vectors and from hosts:
\begin{equation}
\begin{aligned}
N_h &= S_h(0) + E_h(0) + I_{h1}(0) + I_{h2}(0) + A_h(0) + R_h(0) = 0.518628,\\
N_v &= 5 N_h,\\
F_v &= ab\,\frac{I_v}{N_h} S_h,\\
F_h &= \beta\,\frac{\kappa E_h + I_{h1} + \tau I_{h2}}{N_h} S_h.
\end{aligned}\label{eq:app:zika:aux}
\end{equation}
The equations, as printed in \citet{Shaier2022DINN} and used by the generator, are
\begin{equation}
\begin{aligned}
\dot S_h &= -F_v - F_h,\\
\dot E_h &= -\theta(F_v + F_h) - V_h E_h,\\
\dot I_{h1} &= V_h E_h - \gamma_{h1} I_{h1},\\
\dot I_{h2} &= \gamma_{h1} I_{h1} - \gamma_{h2} I_{h2},\\
\dot A_h &= (1 - \theta)(F_v - F_h) - \gamma_h A_h,\\
\dot R_h &= \gamma_{h2} I_{h2} + \gamma_h A_h,\\
\dot S_v &= \mu_v N_v - ac\,\frac{\eta E_h + I_{h1}}{N_h} S_v - \mu_v S_v,\\
\dot E_v &= ac\,\frac{\eta E_h + I_{h1}}{N_h} - (V_v + \mu_v) E_v,\\
\dot I_v &= V_v E_v - \mu_v I_v.
\end{aligned}\label{eq:app:zika}
\end{equation}
Parameters, coefficients and initial condition:
\begin{equation}
\begin{aligned}
\bm\omega &= (b, c, \eta, \beta, \kappa, \tau, \theta, V_h, V_v, \gamma_{h1}, \gamma_{h2}, \gamma_h, \mu_v) \in \R^{13},\\
\bm\omega^\star &= (0.4,\ 0.5,\ 0.1,\ 0.05,\ 0.6,\ 0.3,\ 18,\ 1/5,\ 10,\ 1/5,\ 1/20,\ 1/7,\ 1/14),\\
\bm\theta &= g(\bm\omega) = (ab,\ \beta\kappa,\ \beta,\ \beta\tau,\ \theta ab,\ \theta\beta\kappa,\ \theta\beta,\ \theta\beta\tau,\ V_h,\ \gamma_{h1},\ \gamma_{h2},\ \gamma_h,\ \mu_v,\ ac\eta,\ ac,\ V_v) \in \R^{16},\\
\bm\omega &= h(\bm\theta) = (\vartheta_1/a,\ \vartheta_{15}/a,\ \vartheta_{14}/\vartheta_{15},\ \vartheta_3,\ \vartheta_2/\vartheta_3,\ \vartheta_4/\vartheta_3,\ \vartheta_7/\vartheta_3,\ \vartheta_9,\ \vartheta_{16},\ \vartheta_{10},\ \vartheta_{11},\ \vartheta_{12},\ \vartheta_{13}),\\
\bx_0 &= (0.516,\ 6.57\cdot 10^{-4},\ 6.57\cdot 10^{-4},\ 6.57\cdot 10^{-4},\ 6.57\cdot 10^{-4},\ 0,\ 10^{-4},\ 10^{-4},\ 10^{-4}).
\end{aligned}
\end{equation}
The three constraints are $\vartheta_5 = \vartheta_7\vartheta_1/\vartheta_3$, $\vartheta_6 = \vartheta_7\vartheta_2/\vartheta_3$ and
$\vartheta_8 = \vartheta_7\vartheta_4/\vartheta_3$. The biting rate $a$ enters only through $ab$ and $ac$ and is therefore
fixed. The parameter $m$ of the DINN table does not occur in the model; the generator uses $\gamma_{h2} = 1/20$.
Data: $m = 50$ samples on $[0, 30]$ days.
\begin{table}[htbp]\centering\scriptsize
\caption{Zika: PIWR estimates and PIWR-RACC 90\,\% simultaneous intervals, and the DINN values are from \citet{Shaier2022DINN}.}
\label{tab:app:T11_zika}
\setlength{\tabcolsep}{3pt}
\begin{tabular}{l r r r r r c r r c}
\toprule
 & & \multicolumn{2}{c}{DINN} & \multicolumn{3}{c}{PIWR / PIWR-RACC, $\epsilon = 0$} & \multicolumn{3}{c}{PIWR / PIWR-RACC, $\epsilon = 0.01$}\\
\cmidrule(lr){3-4}\cmidrule(lr){5-7}\cmidrule(lr){8-10}
Par. & Actual & Found & \% err. & Found & \% err. & 90\,\% interval & Found & \% err. & 90\,\% interval\\
\midrule
$b$ & 0.4 & 0.4033 & 0.83 & 0.5677 & 41.9 & [-39.45, 40.59] & -1.459 & 465 & [-125.4, 122.5] \\
$c$ & 0.5 & 0.5015 & 0.309 & 0.4683 & 6.33 & [0.416, 0.5207] & 0.4627 & 7.47 & [0.2723, 0.653] \\
$\eta$ & 0.1 & 0.0999 & 0.0687 & 0.09803 & 1.97 & [0.08887, 0.1072] & 0.09615 & 3.85 & [0.06212, 0.1302] \\
$\beta$ & 0.05 & 0.0498 & 0.41 & -0.07811 & 256 & [-8.387, 8.23] & 0.1867 & 273 & [-25.54, 25.92] \\
$\kappa$ & 0.6 & 0.6033 & 0.549 & -1.354 & 326 & [-83.54, 80.84] & -0.3063 & 151 & [-39.06, 38.45] \\
$\tau$ & 0.3 & 0.2902 & 3.26 & -0.4845 & 261 & [-34.35, 33.38] & -0.1018 & 134 & [-9.242, 9.039] \\
$\theta$ & 18 & 17.97 & 0.184 & -10.48 & 158 & [-1149, 1128] & 6.272 & 65.2 & [-832.9, 845.5] \\
$V_h$ & 0.2 & 0.1996 & 0.18 & 0.2 & 0.00322 & [0.1783, 0.2217] & 0.2007 & 0.326 & [0.1206, 0.2807] \\
$V_v$ & 10 & 10.02 & 0.17 & 9.353 & 6.47 & [8.373, 10.33] & 9.071 & 9.29 & [5.577, 12.56] \\
$\gamma_{h1}$ & 0.2 & 0.1991 & 0.465 & 0.2 & 0.0011 & [0.1407, 0.2593] & 0.1992 & 0.4 & [-0.01941, 0.4178] \\
$\gamma_{h2}$ & 0.05 & 0.0504 & 0.726 & 0.05 & 0.00167 & [0.03633, 0.06367] & 0.04992 & 0.168 & [-0.0008461, 0.1007] \\
$\gamma_h$ & 0.1429 & 0.1406 & 1.6 & 0.1429 & 0.000283 & [0.1225, 0.1632] & 0.1411 & 1.24 & [0.06605, 0.2161] \\
$\mu_v$ & 0.07143 & 0.0723 & 1.18 & 0.07128 & 0.211 & [0.05852, 0.08404] & 0.07144 & 0.0161 & [0.02437, 0.1185] \\
$a$ & 0.5 & 0.4997 & 0.0588 & \multicolumn{6}{l}{known: only $ab$ and $ac$ are identifiable} \\
$m$ & 5 & 5.2937 & 5.87 & \multicolumn{6}{l}{not in the model} \\
\bottomrule
\end{tabular}
\end{table}

\bibliographystyle{unsrtnat-nourl}   

\section{Further benchmark models}\label{app:extra}

\subsection{S3I3R model}\label{app:s3i3r}
The model of \citet[eq.~21]{Nielsen2025PIR} has the state $\bx = (S, I_1, I_2, I_3, R_1, R_2, R_3)$, $n = 7$:
susceptibles $S$, three infected stages $I_1$ (infected), $I_2$ (hospitalized) and $I_3$ (in intensive care),
recovered $R_1$, vaccinated $R_2$ and dead $R_3$. With the population $N$ and the vaccination rate $\tau$ known,
\begin{equation}
\begin{aligned}
\dot S &= -\Big(\frac{\beta I_1}{N} + \frac{\tau}{S + I_1 + R_1}\Big) S,\\
\dot I_1 &= \frac{\beta I_1 S}{N} - \Big(\gamma_1 + \frac{\tau}{S + I_1 + R_1} + \phi_1\Big) I_1,\\
\dot I_2 &= \phi_1 I_1 - (\gamma_2 + \phi_2) I_2,\\
\dot I_3 &= \phi_2 I_2 - (\gamma_3 + \theta) I_3,\\
\dot R_1 &= \gamma_1 I_1 + \gamma_2 I_2 + \gamma_3 I_3 - \frac{\tau R_1}{S + I_1 + R_1},\\
\dot R_2 &= \tau,\\
\dot R_3 &= \theta I_3.
\end{aligned}\label{eq:app:s3i3r}
\end{equation}
The terms with $\tau$ form the offset $f_0$ of \eqref{eq:linear_known}; here $\tau = 0$. With all seven rates unknown,
and with fractional states ($N = 1$),
\begin{equation}
\begin{aligned}
\bm\omega &= (\beta, \gamma_1, \gamma_2, \gamma_3, \phi_1, \phi_2, \theta) \in \R^7,\\
\bm\omega^\star &= (1/2,\ 1/3,\ 1/20,\ 1/20,\ 1/20,\ 1/20,\ 1/10),\\
\bm\theta &= g(\bm\omega) = (\beta, \gamma_1, \gamma_2, \gamma_3, \phi_1, \phi_2, \theta) \in \R^7,\\
\bm\omega &= h(\bm\theta) = (\vartheta_1, \dots, \vartheta_7),\\
\bx_0 &= (0.9999,\ 0.0001,\ 0,\ 0,\ 0,\ 0,\ 0).
\end{aligned}
\end{equation}
In the setting of \citet{Nielsen2025PIR}, the recovery rates are assumed known and hence their terms must join the offset,
\begin{equation}
\begin{aligned}
\bm\omega &= (\beta, \phi_1, \phi_2, \theta) \in \R^4,\\
\bm\theta &= g(\bm\omega) = (\beta, \phi_1, \phi_2, \theta) \in \R^4,\\
f_0(\bx) &= \big(0,\ -\gamma_1 I_1,\ -\gamma_2 I_2,\ -\gamma_3 I_3,\ \gamma_1 I_1 + \gamma_2 I_2 + \gamma_3 I_3,\ 0,\ 0\big).
\end{aligned}
\end{equation}
The setting uses the same initial condition. These values are taken from the code of \citet{Nielsen2025PIR}; the parameter vector printed in their eq.~(26) lacks
one of the entries $1/20$. Data: $m = 29$ daily samples on $[0, 28]$ days.

\subsection{SIR model with only the infected observed}\label{app:school}
The SIR model with the state $(S, I, R)$ and known population $N = 763$ is
\begin{equation}
\begin{aligned}
\dot S &= -\beta\frac{S I}{N},\\
\dot I &= \beta\frac{S I}{N} - \gamma I,\\
\dot R &= \gamma I.
\end{aligned}\label{eq:app:sir}
\end{equation}
It conserves $S + I + R = N$. Parameters, coefficients and initial condition:
\begin{equation}
\begin{aligned}
\bm\omega &= (\beta, \gamma) \in \R^2,\\
\bm\theta &= g(\bm\omega) = (\beta - \gamma,\ \beta/N,\ \beta\gamma/N) \in \R^3,\\
\bm\omega &= h(\bm\theta) = (N\vartheta_2,\ \vartheta_3/\vartheta_2),\\
\mathcal R_0 &= \beta/\gamma = N\vartheta_2^2/\vartheta_3,\\
(S(0), I(0), R(0)) &= (762,\ 1,\ 0).
\end{aligned}
\end{equation}
Only $I$ is observed, so the regression uses the scalar state $\bx = I$, $n = 1$. With $R = \gamma C$ and
$S = N - I - \gamma C$, the infected compartment satisfies the closed equation \eqref{eq:school},
\begin{equation}
\begin{aligned}
\dot I = (\beta - \gamma) I - \frac{\beta}{N} I^2 - \frac{\beta\gamma}{N} I\,C(t), \quad C(t) &= \int_0^t I(s)\,ds,
\end{aligned}
\end{equation}
whose library $A = (I,\ -I^2,\ -I\,C)$ depends on the past of the trajectory through $C$, which is computed from
the data by the cubic-spline quadrature of \cref{sec:piwr}. The one constraint is
$\vartheta_3 = \vartheta_2(N\vartheta_2 - \vartheta_1)$; PIWR-VP imposes it (\cref{sec:res:extra}).
There are no true values $\bm\omega^\star$ for real data. The data (\texttt{influenza\_england\_1978\_school} in the
R package \texttt{outbreaks}) are the number of boys in bed on 22~January to 4~February 1978, placed at
$t = 1, \dots, 14$ days, together with $I(0) = 1$ at $t = 0$ as in \citet{Grinsztajn2021Stan}:
$3, 8, 26, 76, 225, 298, 258, 233, 189, 128, 68, 29, 14, 4$.

\bibliography{references}

\begin{thebibliography}{27}
\providecommand{\natexlab}[1]{#1}
\providecommand{\url}[1]{\texttt{#1}}
\expandafter\ifx\csname urlstyle\endcsname\relax
  \providecommand{\doi}[1]{doi: #1}\else
  \providecommand{\doi}{doi: \begingroup \urlstyle{rm}\Url}\fi

\bibitem[Raissi et~al.(2019)Raissi, Perdikaris, and Karniadakis]{RAISSI2019686}
M.~Raissi, P.~Perdikaris, and G.~E. Karniadakis.
\newblock Physics-informed neural networks: A deep learning framework for
  solving forward and inverse problems involving nonlinear partial differential
  equations.
\newblock \emph{Journal of Computational Physics}, 378:\penalty0 686--707,
  2019.
\newblock \doi{10.1016/j.jcp.2018.10.045}.

\bibitem[Karniadakis et~al.(2021)Karniadakis, Kevrekidis, Lu, Perdikaris, Wang,
  and Yang]{karniadakis2021physics}
G.~E. Karniadakis, I.~G. Kevrekidis, L.~Lu, P.~Perdikaris, S.~Wang, and
  L.~Yang.
\newblock Physics-informed machine learning.
\newblock \emph{Nature Reviews Physics}, 3\penalty0 (6):\penalty0 422--440,
  2021.
\newblock \doi{10.1038/s42254-021-00314-5}.

\bibitem[Cuomo et~al.(2022)Cuomo, Di~Cola, Giampaolo, Rozza, Raissi, and
  Piccialli]{cuomo2022scientific}
S.~Cuomo, V.~S. Di~Cola, F.~Giampaolo, G.~Rozza, M.~Raissi, and F.~Piccialli.
\newblock Scientific machine learning through physics--informed neural
  networks: Where we are and what’s next.
\newblock \emph{Journal of Scientific Computing}, 92\penalty0 (3):\penalty0 88,
  2022.
\newblock \doi{10.1007/s10915-022-01939-z}.

\bibitem[Shaier et~al.(2022)Shaier, Raissi, and Seshaiyer]{Shaier2022DINN}
S.~Shaier, M.~Raissi, and P.~Seshaiyer.
\newblock Data-driven approaches for predicting spread of infectious diseases
  through {DINNs}: Disease informed neural networks.
\newblock \emph{Letters in Biomathematics}, 8\penalty0 (1):\penalty0 71--105,
  2022.
\newblock arXiv:2110.05445.

\bibitem[Nielsen et~al.(2025)Nielsen, Jacobsen, Olson, S{\o}rensen, and
  Engsig-Karup]{Nielsen2025PIR}
J.~S. Nielsen, M.~G. Jacobsen, A.~B. Olson, M.~P. S{\o}rensen, and A.~P.
  Engsig-Karup.
\newblock Physics-informed regression: Parameter estimation in parameter-linear
  nonlinear dynamic models.
\newblock Preprint: arXiv:2508.19249, 2025.

\bibitem[Bates and Watts(1988)]{BatesWatts1988}
D.~M. Bates and D.~G. Watts.
\newblock \emph{Nonlinear Regression Analysis and Its Applications}.
\newblock Wiley, New York, 1988.
\newblock \doi{10.1002/9780470316757}.

\bibitem[Raue et~al.(2009)Raue, Kreutz, Maiwald, Bachmann, Schilling,
  Klingm{\"u}ller, and Timmer]{Raue2009Profile}
A.~Raue, C.~Kreutz, T.~Maiwald, J.~Bachmann, M.~Schilling, U.~Klingm{\"u}ller,
  and J.~Timmer.
\newblock Structural and practical identifiability analysis of partially
  observed dynamical models by exploiting the profile likelihood.
\newblock \emph{Bioinformatics}, 25\penalty0 (15):\penalty0 1923--1929, 2009.
\newblock \doi{10.1093/bioinformatics/btp358}.

\bibitem[Efron(1979)]{Efron1979Bootstrap}
B.~Efron.
\newblock Bootstrap methods: another look at the jackknife.
\newblock \emph{The Annals of Statistics}, 7\penalty0 (1):\penalty0 1--26,
  1979.
\newblock \doi{10.1214/aos/1176344552}.

\bibitem[Chowell(2017)]{Chowell2017Primer}
G.~Chowell.
\newblock Fitting dynamic models to epidemic outbreaks with quantified
  uncertainty: a primer for parameter uncertainty, identifiability, and
  forecasts.
\newblock \emph{Infectious Disease Modelling}, 2\penalty0 (3):\penalty0
  379--398, 2017.
\newblock \doi{10.1016/j.idm.2017.08.001}.

\bibitem[Carpenter et~al.(2017)Carpenter, Gelman, Hoffman, Lee, Goodrich,
  Betancourt, Brubaker, Guo, Li, and Riddell]{Carpenter2017Stan}
B.~Carpenter, A.~Gelman, M.~D. Hoffman, D.~Lee, B.~Goodrich, M.~Betancourt,
  M.~Brubaker, J.~Guo, P.~Li, and A.~Riddell.
\newblock Stan: a probabilistic programming language.
\newblock \emph{Journal of Statistical Software}, 76\penalty0 (1):\penalty0
  1--32, 2017.
\newblock \doi{10.18637/jss.v076.i01}.

\bibitem[Grinsztajn et~al.(2021)Grinsztajn, Semenova, Margossian, and
  Riou]{Grinsztajn2021Stan}
L.~Grinsztajn, E.~Semenova, C.~C. Margossian, and J.~Riou.
\newblock Bayesian workflow for disease transmission modeling in {S}tan.
\newblock \emph{Statistics in Medicine}, 40\penalty0 (27):\penalty0 6209--6234,
  2021.
\newblock \doi{10.1002/sim.9164}.

\bibitem[Toni et~al.(2009)Toni, Welch, Strelkowa, Ipsen, and
  Stumpf]{Toni2009ABC}
T.~Toni, D.~Welch, N.~Strelkowa, A.~Ipsen, and M.~P.~H. Stumpf.
\newblock Approximate {B}ayesian computation scheme for parameter inference and
  model selection in dynamical systems.
\newblock \emph{Journal of the Royal Society Interface}, 6\penalty0
  (31):\penalty0 187--202, 2009.
\newblock \doi{10.1098/rsif.2008.0172}.

\bibitem[Yang et~al.(2021)Yang, Meng, and Karniadakis]{yang2021bpinn}
L.~Yang, X.~Meng, and G.~E. Karniadakis.
\newblock {B-PINNs}: {B}ayesian physics-informed neural networks for forward
  and inverse {PDE} problems with noisy data.
\newblock \emph{Journal of Computational Physics}, 425:\penalty0 109913, 2021.
\newblock \doi{10.1016/j.jcp.2020.109913}.

\bibitem[Kennedy and O'Hagan(2001)]{Kennedy2001Calibration}
M.~C. Kennedy and A.~O'Hagan.
\newblock Bayesian calibration of computer models.
\newblock \emph{Journal of the Royal Statistical Society: Series B (Statistical
  Methodology)}, 63\penalty0 (3):\penalty0 425--464, 2001.
\newblock \doi{10.1111/1467-9868.00294}.

\bibitem[Jensen et~al.(2022)Jensen, Engsig-Karup, and Knudsen]{Jensen2022PC}
B.~C.~S. Jensen, A.~P. Engsig-Karup, and K.~Knudsen.
\newblock Efficient uncertainty quantification and variance-based sensitivity
  analysis in epidemic modelling using polynomial chaos.
\newblock \emph{Mathematical Modelling of Natural Phenomena}, 17:\penalty0 8,
  2022.
\newblock \doi{10.1051/mmnp/2022014}.

\bibitem[Vovk et~al.(2005)Vovk, Gammerman, and Shafer]{vovk2005algorithmic}
V.~Vovk, A.~Gammerman, and G.~Shafer.
\newblock \emph{Algorithmic Learning in a Random World}.
\newblock Springer, New York, 2005.
\newblock \doi{10.1007/b106715}.

\bibitem[Shafer and Vovk(2008)]{shafer2008tutorial}
G.~Shafer and V.~Vovk.
\newblock A tutorial on conformal prediction.
\newblock \emph{Journal of Machine Learning Research}, 9\penalty0
  (12):\penalty0 371--421, 2008.

\bibitem[Lei et~al.(2018)Lei, G'Sell, Rinaldo, Tibshirani, and
  Wasserman]{lei2018distribution}
J.~Lei, M.~G'Sell, A.~Rinaldo, R.~J. Tibshirani, and L.~Wasserman.
\newblock Distribution-free predictive inference for regression.
\newblock \emph{Journal of the American Statistical Association}, 113\penalty0
  (523):\penalty0 1094--1111, 2018.
\newblock \doi{10.1080/01621459.2017.1307116}.

\bibitem[Angelopoulos and Bates(2023)]{angelopoulos2023gentle}
A.~N. Angelopoulos and S.~Bates.
\newblock Conformal prediction: A gentle introduction.
\newblock \emph{Foundations and Trends in Machine Learning}, 16\penalty0
  (4):\penalty0 494--591, 2023.
\newblock \doi{10.1561/2200000101}.

\bibitem[Varah(1982)]{Varah1982}
J.~M. Varah.
\newblock A spline least squares method for numerical parameter estimation in
  differential equations.
\newblock \emph{SIAM Journal on Scientific and Statistical Computing},
  3\penalty0 (1):\penalty0 28--46, 1982.
\newblock \doi{10.1137/0903003}.

\bibitem[Ramsay et~al.(2007)Ramsay, Hooker, Campbell, and Cao]{Ramsay2007}
J.~O. Ramsay, G.~Hooker, D.~Campbell, and J.~Cao.
\newblock Parameter estimation for differential equations: a generalized
  smoothing approach.
\newblock \emph{Journal of the Royal Statistical Society: Series B (Statistical
  Methodology)}, 69\penalty0 (5):\penalty0 741--796, 2007.
\newblock \doi{10.1111/j.1467-9868.2007.00610.x}.

\bibitem[Messenger and Bortz(2021)]{Messenger2021WSINDy}
D.~A. Messenger and D.~M. Bortz.
\newblock Weak {SINDy}: Galerkin-based data-driven model selection.
\newblock \emph{Multiscale Modeling \& Simulation}, 19\penalty0 (3):\penalty0
  1474--1497, 2021.
\newblock \doi{10.1137/20M1343166}.

\bibitem[{Anonymous}(1978)]{BMJ1978Influenza}
{Anonymous}.
\newblock Influenza in a boarding school.
\newblock \emph{British Medical Journal}, 1\penalty0 (6112):\penalty0 587,
  1978.

\bibitem[Millevoi et~al.(2024)Millevoi, Pasetto, and
  Ferronato]{Millevoi2024PINN}
C.~Millevoi, D.~Pasetto, and M.~Ferronato.
\newblock A physics-informed neural network approach for compartmental
  epidemiological models.
\newblock \emph{PLoS Computational Biology}, 20\penalty0 (9):\penalty0
  e1012387, 2024.
\newblock \doi{10.1371/journal.pcbi.1012387}.

\bibitem[Golub and Pereyra(1973)]{Golub1973VarPro}
G.~H. Golub and V.~Pereyra.
\newblock The differentiation of pseudo-inverses and nonlinear least squares
  problems whose variables separate.
\newblock \emph{SIAM Journal on Numerical Analysis}, 10\penalty0 (2):\penalty0
  413--432, 1973.
\newblock \doi{10.1137/0710036}.

\bibitem[Eilers and Marx(1996)]{Eilers1996PSpline}
P.~H.~C. Eilers and B.~D. Marx.
\newblock Flexible smoothing with {B}-splines and penalties.
\newblock \emph{Statistical Science}, 11\penalty0 (2):\penalty0 89--121, 1996.
\newblock \doi{10.1214/ss/1038425655}.

\bibitem[Petzold(1983)]{Petzold1983LSODA}
L.~Petzold.
\newblock Automatic selection of methods for solving stiff and nonstiff systems
  of ordinary differential equations.
\newblock \emph{SIAM Journal on Scientific and Statistical Computing},
  4\penalty0 (1):\penalty0 136--148, 1983.
\newblock \doi{10.1137/0904010}.

\end{thebibliography}

\end{document}